\documentclass[journal]{IEEEtran}
\usepackage{amsmath,amsfonts,amssymb,amsthm,bm}
\usepackage[caption=false,font=normalsize,labelfont=sf,textfont=sf]{subfig}
\usepackage[hyphens]{url}
\usepackage{graphicx}
\usepackage{cite}
\usepackage[colorlinks,citecolor=blue]{hyperref}
\usepackage{physics}
\usepackage{nicefrac}
\usepackage{booktabs}
\usepackage[inline]{enumitem}
\usepackage[nameinlink]{cleveref}
\usepackage[store-sets-label]{keytheorems}
\usepackage{etoc}

\crefname{section}{\S}{\S\S}
\crefformat{section}{#2§#1#3}

\newkeytheorem{theorem,lemma,corollary,definition}

\DeclareMathOperator*{\argmin}{arg\,min}

\makeatletter
\def\@IEEEsectpunct{.\ \,}
\def\paragraph{\@startsection{paragraph}{4}{\z@}{1.5ex plus 1.5ex minus 0.5ex}%
{0ex}{\bfseries}}
\makeatother

\begin{document}

\etocdepthtag.toc{main}
\etocsettagdepth{main}{subsubsection}
\etocsettagdepth{appendix}{none}

\title{Understanding Rate-Distortion Performance in Distributed Transformer Inference}

\author{
Anderson de Andrade,~\IEEEmembership{Student Member,~IEEE,}
Alon Harell,~\IEEEmembership{Student Member,~IEEE,}
Ivan V. Baji\'{c},~\IEEEmembership{Senior Member,~IEEE}% <-this % stops a space
\thanks{Authors are with Simon Fraser University, 8888 University Drive, Burnaby, BC, V5S 1A6, Canada. Email communications to Anderson de Andrade: \texttt{anderson\_de\_andrade@sfu.ca}.}% <-this % stops a space
\thanks{This work was partially funded by Intel Labs and the Natural Sciences and Engineering Research Council of Canada (NSERC).}% <-this % stops a space
\thanks{This work has been submitted to the IEEE for possible publication. Copyright may be transferred without notice, after which this version may no longer be accessible.}%
}

% The paper headers
\markboth{Submitted for review}%
{de Andrade \MakeLowercase{\textit{et al.}}: Understanding Rate-Distortion Performance in Distributed Transformer Inference}

\maketitle
\IEEEpeerreviewmaketitle

\begin{abstract}
Transformers achieve superior performance on many tasks, but impose heavy compute and memory requirements during inference. This inference can be made more efficient by partitioning the process across multiple devices, which, in turn, requires compressing its intermediate representations. We study compressibility of transformer's intermediate representations via learned compression through the lens of rate-distortion and the theory of usable information. Our study reveals that, unlike convolutional models, deeper representations in transformers become more difficult to compress. The reason for this behavior is twofold: first, the complexity of representation increases as we move deeper into the transformer, necessitating a higher rate; and second, the higher complexity of representations worsens the generaliztion bound for learned entropy estimates, which further compromises compression performance. Through a combination of  experiments and theory, we characterize and analyze the compressibility of transformer representations, derive bounds on the achievable rate of learned codecs applied to these representations, and offer a unified lens for understanding rate-distortion performance in representation coding. 
\end{abstract}


\section{Introduction}

\IEEEPARstart{T}{ransformers}~\cite{Vaswani2017AttentionIA} have become predominant in many machine learning tasks, such as language~\cite{GPT2,biderman2023pythia} and vision-language modeling~\cite{Zhang2024TPAMI}, image recognition~\cite{DosovitskiyB0WZ21}, and representation learning of unstructured data~\cite{Nakhli2023SparseMG}. These models often have billions of parameters~\cite{GPT2, biderman2023pythia,Zhang2024TPAMI}, demanding an immense amount of computational resources. One way to accommodate such an immense model size is to distribute the computation among multiple devices.  Splitting the transformer architecture into manageable modules of contiguous layers allows better horizontal scaling across multiple heterogeneous devices~\cite{MarufAAS24, JafariSZWZ24}.
For example, in the Internet of Things (IoT) setting~\cite{Shlezinger2022,MatsubaraLR23, KoJJP24}, a mobile device could perform part of the inference process and then transmit the intermediate representation to a cloud or data center to complete the inference.

\begin{figure}[!t]
    \centering
    \subfloat[Distribution across time steps]{
        \includegraphics[width=0.86\columnwidth]{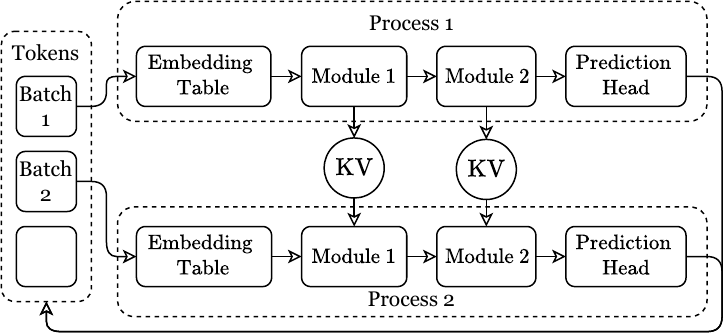}
    }
    \hfil
    \subfloat[Distribution across modules]{
        \includegraphics[width=0.97\columnwidth]{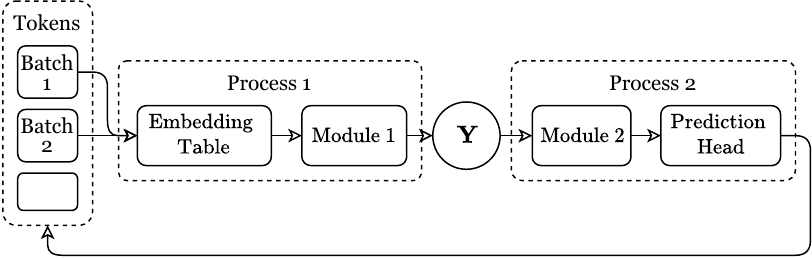}
    }
    \caption{Two scenarios of distributed transformer inference: (a) Same module on multiple devices, inferring different time steps, requiring the transmission of the KV cache; (b) Different modules on different devices, requiring transmission of intermediate representations $\mathbf{Y}$. Our focus is the latter scenario.}
    \label{fig:inference}
\end{figure}

There are several ways in which the transformer inference process can be divided among multiple devices. The same module -- a contiguous subset of operations -- could be deployed on multiple devices to perform inference for different time steps, as shown in Fig.~\ref{fig:inference}(a). In such situations, the \textit{KV cache}~\cite{Vaswani2017AttentionIA} should be transmitted along with the intermediate representations. Alternatively, different modules could be deployed on different devices, as shown in Fig.~\ref{fig:inference}(b). In this setting, which is the focus of this paper, only the corresponding intermediate representations must be transmitted.

Intermediate representations are usually compressed~\cite{Choi2018ICIP} in order to conserve bandwidth, reduce transmission latency, and allow a larger number of users to share the communication infrastructure. In this paper, we study compressibility of intermediate representations produced by transformer modules and the effect of lossy compression on inference accuracy.
In particular, we construct several learning-based compression strategies -- including learned entropy models -- following established principles from learned image compression~\cite{balle2018variational}. This allows adaptation to the statistics of transformer's representations and makes rate-distortion (RD) optimization feasible. Some of these compression strategies are able to achieve two orders of magnitude compression without affecting the performance in terms of transformer's output perplexity.

Our experiments also reveal an interesting phenomenon: \textit{compression of intermediate representations seems to become more difficult as we move deeper into the transformer}. At first glance, this seems at odds with both the Data Processing Inequality from information theory~\cite{Cover} and the earlier results on compression of intermediate features of Convolutional Neural Networks (CNNs)~\cite{Choi2018ICIP,Choi2022ScalableIC}. Much of the paper is devoted to explaining this phenomenon, as it holds important lessons for the design of transformer-based distributed inference systems.
Specifically, we explain this behavior through the theory of \textit{usable information}~\cite{XuZSSE20}, which takes into account the modeling power and computational constraints of an entropy model. We derive different bounds on $\mathcal{V}$-entropy~\cite{XuZSSE20}
and provide a generalization error bound in terms of the \textit{Rademacher complexity} \cite{LearningTheory} of the target representation, and the \textit{Lipschitz constant} \cite{benyaminigeometric} of the entropy model.
The bounds on $\mathcal{V}$-entropy explain why, due to the
limitations of the entropy model, the actual achieved rate is often orders of magnitude higher than the entropy of the target representation. Moreover, the bounds on its generalization error explain why an increase in the complexity of the entropy model does not necessarily translate into better compression performance.

As contributions:
\begin{enumerate*}
    \item we bring rate-distortion optimization for split inference to transformers and language tasks, achieving significant compression rates that offer a promising avenue for token compression (\cref{sec:entropy-models,sec:rd-benchmarks});
    \item we provide a rate characterization of the intermediate representations of transformers, demonstrating an unexpected decrease in rate-distortion performance as we code deeper representations, and identifying
    reasons behind such behavior (\cref{sec:theory,sec:rd-split-exp});
    \item we extend the theory of usable information, providing further insights that are relevant to learned compression in general, and compression of transformer representations in particular (\cref{sec:theory}, Supplementary Material).
\end{enumerate*}

\section{Related work} 
Several techniques for making transformer inference more efficient have been studied in the context of large language models~\cite{LLMSurvey}. Such approaches include request batching and scheduling~\cite{YuJKKC22, AgrawalKPMKGTR24, FlexGen, FastDecode, PatelCZSGMB24}, parameter pruning~\cite{Zhang0SYOYZ24}, model quantization~\cite{0002TTYCWXDG024}, token pruning~\cite{JiangWLYQ23}, and sparse attention mechanisms~\cite{Longformer}. The approaches in~\cite{FlexGen, FastDecode, PatelCZSGMB24} share computation between multiple processes, but none consider optimizing the intermediate representations in terms of rate-distortion.

In this work, we use techniques from learned image compression~\cite{balle2018variational, MinnenBT18, ZouSZ22, JiangYZGW25}, to enable rate-distortion optimization of distributed transformer inference. A typical learned image codec consists of an \textit{analysis transform} that maps an image into a quantized latent representation; an \textit{entropy encoder}, which incorporates an \textit{entropy model} of the latent representation to produce an efficient binary code; an \textit{entropy decoder}, which uses the same entropy model and maps the binary code back to the quantized latent representation; and a \textit{synthesis transform} to convert the latent representation into an approximation of the input image. Seminal work on learned image compression employed CNNs~\cite{balle2018variational, MinnenBT18, HeYPMQW22,JiangYZNGW23}. More recent work uses transformer architectures~\cite{ZouSZ22,LiLDLZX24} and state-space models~\cite{MambaVC,MambaIC}. 

Various entropy models have been proposed for learned compression. Often, a hyper-prior~\cite{balle2018variational} is involved in modeling the distribution of the latent representation and is coded as side information.
The latent representation is often trained to be, and modeled as, a multivariate normal with a diagonal covariance matrix, while the distribution imposed on the hyper-prior is fully-factorized and non-parametric~\cite{balle2018variational}. In~\cite{FuenteSB24}, a Fourier basis is proposed to model a fully-factorized probability distribution. Using fewer parameters, this approach is able to fit more complex distributions. In this work, we use several hyper-prior-based entropy models for the representations produced by transformed modules.

Another related area of work is \textit{coding for machines} (CfM)~\cite{Choi2022ScalableIC, HarellFADKSTAB25}, where the goal is to create a compressed reresentation of the input signal -- usually image or video -- for the purpose of performing inference from that compressed representation. Compression of intermediate representations is well-established within CfM~\cite{Choi2018ICIP}, and has led to the emerging compression standard on feature coding for machines.\footnote{ISO/IEC CD 23888-4: Information technology — Artificial intelligence for multimedia, 
Part 4: Feature coding for machines} However, most of the prior work on CfM and the emerging standard focus on compressing convolutional features. Compression of transformer's intermediate representations is the next big challenge, and the present paper contributes to the understanding of the various tradeoffs involved in it.

In CfM, the rate of an \textit{optimal} codec for perfect inference is lower-bounded by the entropy of the target random variable, and can be even lower when inference is imperfect~\cite{bajić2025rateaccuracyboundsvisualcoding}. While existing CfM codecs offer substantial rate reductions~\cite{AndradeB24,HarellFADKSTAB25} compared to conventional image or video codecs, the rates they achieve in many cases are orders of magnitude higher than the theoretical rate-accuracy bounds~\cite{bajić2025rateaccuracyboundsvisualcoding}. We show that the ability of a learned codec to achieve optimality is limited by the complexity of both the codec and its optimization. The lowest \textit{achievable} entropy a set of probability functions $\mathcal{V}$ can measure in a random variable is formalized by the concept of \textit{$\mathcal{V}$-entropy} \cite{XuZSSE20}. It can be estimated with guarantees if the richness of $\mathcal{V}$ is bounded in terms of its \textit{Rademacher complexity} \cite{LearningTheory}. It was shown in \cite{XuZSSE20} that bounds on the complexity of $\mathcal{V}$ directly translate to \textit{probably approximately correct} (PAC) \cite{Valiant84} bounds for $\mathcal{V}$-entropy estimation. We use these prior results to shed light on the challenges involved in compressing transformer's intermediate representations.

\section{Preliminaries}

\begin{figure*}[!t]
    \includegraphics[width=\linewidth]{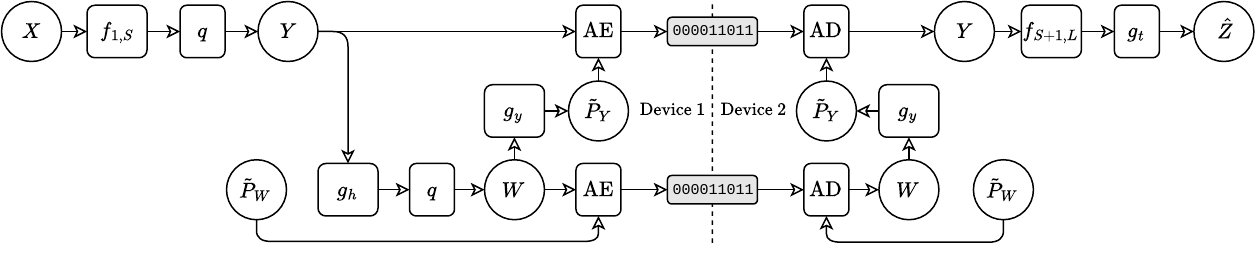}
    \caption{Overview of the system used to study compressibility of transformer's intermediate representation $Y$. The AE and AD blocks correspond to arithmetic encoders and decoders, respectively. They use the probability distributions provided by the entropy models to encode their target representation into a bitstream and decode it back. The dotted line separates two devices, with bitstreams (gray blocks) connecting them.}
    \label{fig:codec}
\end{figure*}

For concreteness and tractability, in this work we focus on decoder-only transformers \cite{LiuSPGSKS18,Radford2018ImprovingLU} split into two modules, with the goal of compressing a single  intermediate representation passed on from the first module to the second one. Findings obtained in this setting also apply to a larger number of modules and to encoder-decoder transformer architectures. A high-level overview of the system is shown in Fig.~\ref{fig:codec}.  To simplify notation without loss of generality, we assume that all transformer layers produce representations of the same dimensionality as the input embeddings. See the Supplementary Material for a summary of the notation used.

Let $X \in \mathbb{R}^{T \times E}$ be an input random variable, where $T$ is the size of a time or spatial dimension, and $E$ is the embedding size. Let $\{f_l : \mathbb{R}^{T \times E} \to \mathbb{R}^{T \times E} \}_{l=1}^{L}$ be a set of $L$ transformer blocks describing the bulk of a transformer-based neural network. The first module of the network produces the intermediate representation as:
\begin{align}
    Y
    =
    (q \circ f_{1,S})(X); f_{1,S} = f_S \circ ... \circ f_1
    ,
\end{align}
where $S$ is the \textit{split point}, $q$ is a \textit{quantization function} \cite{AgustssonT20, TheisSCH17, BalleLS16}, and $\circ$ denotes the \textit{composition operator}. This quantization function discretizes the target representation to enable efficient coding. It has a differentiable training-time approximation that allows gradient propagation during automatic differentiation. The second module of the network produces the predictions as:
\begin{align}
    \hat{Z}
    =
    (g_t \circ f_{S+1,L})(Y); f_{S+1,L} = f_L \circ ... \circ f_{S+1}
    ,
\end{align}
where $\smash{g_t : \mathbb{R}^{T \times E} \to \mathcal{Z}}$ are the header layers mapping to the sample space of the target random variable $Z$. 

A hyper-prior \cite{balle2018variational} is a random variable $W$ with sample space $\mathcal{W} \subseteq \smash{\mathbb{Z}^{T \times C}}$. An entropy model assumes a learned continuous cumulative distribution function (CDF) for each of the $C$ dimensions of the hyper-prior. Using this entropy model, the rate $\smash{r_w(\mathbf{w}): \mathcal{W} \to \mathbb{R}}$ of a hyper-prior $\mathbf{w}$ is the fully-factorized negative log-likelihood of a unit interval centered around $\mathbf{w}$ \cite{balle2018variational}. See the Supplementary Material for a more detailed definition.

Let $\Omega = \{\mathcal{W} \cup \{ \oslash \} \to \mathcal{P}(\mathcal{Y}) \}$ be the set of functions that map the hyper-prior $W$ or a constant $\oslash$ to any probability distribution over the sample space of $Y$. A \textit{predictive family}~\cite{XuZSSE20} $\mathcal{V} \subseteq \Omega$ is the set of predictive models a learning algorithm is allowed to use due to computational or other constraints. %, or \textit{rate constraints}. 
The predictive conditional $\mathcal{V}$-entropy is defined as:
\begin{align}
\label{eq:conditional-entropy}
H_{\mathcal{V}}(Y|W)
&\triangleq
\inf_{g \in \mathcal{V}}
\mathbb{E}_{\mathbf{w},\mathbf{y} \sim W,Y}
\left[
- \log g[\mathbf{w}](\mathbf{y})
\right]
.
\end{align}
Setting $\mathcal{V} = \Omega$ recovers the usual conditional entropy from information theory~\cite{Cover}. In addition, setting $\mathcal{W} = \oslash$ (i.e., no side information) recovers the unconditional (Shannon's) entropy~\cite{Cover}.

In practice, $\mathcal{V}$-entropy is usually minimized on a dataset, since the true joint distribution $P_{Y,W}$ of the latent representation $Y$ and side information $W$ is not known. Learning theory~\cite{LearningTheory} establishes the generalization error as the discrepancy between the in-sample error (empirical risk) and the expectation of the out-of-sample error.  
This quantity can usually be bounded by the complexity of the hypothesis family (model architecture). Intuitively, a hypothesis family with a smaller complexity is easier to learn. An analog result exists for $\mathcal{V}$-entropy, where, for a set of samples $\mathcal{D} = \{(\mathbf{y}_i, \mathbf{w}_i) \}_{i=1}^N \sim Y, W$, the bound is placed on:
\begin{align}
    \label{eq:v-ent-gen-error}
    R_{\mathcal{V},\mathcal{D}}(Y|W)
    \!
    \triangleq
    \!
    \Bigg\vert
    H_{\mathcal{V}}(Y|W)
    \!
    -
    \!
    \inf_{g \in \mathcal{V}}
    \frac{1}{N}
    \!\!
    \sum_{\mathbf{y},\mathbf{w} \in \mathcal{D}}
    \!\!\!\!\!
    -
    \log
    g[\mathbf{w}](\mathbf{y})
    \Bigg\vert
    .
\end{align}
In a result presented in \cite{XuZSSE20}, the complexity of the predictive family $\mathcal{V}$ emerges in a probably approximately correct (PAC) \cite{LearningTheory} bound for $\mathcal{V}$-entropy estimation. We use these concepts to explain some of the phenomena related to compressibility of transformer's intermediate representations. 

\section{Entropy modeling in transformers}
\label{sec:entropy-models}

\begin{figure*}[!t]
    \subfloat[Standalone hyper-prior]{
        \includegraphics[width=0.42\linewidth]{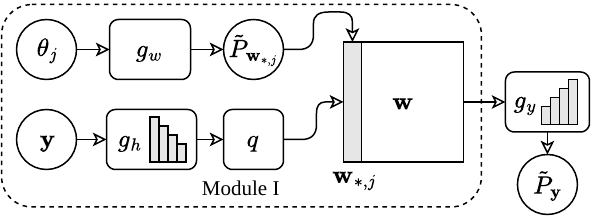}
        \label{fig:proposed}
    }
    \subfloat[Direct access]{
        \includegraphics[width=0.55\linewidth]{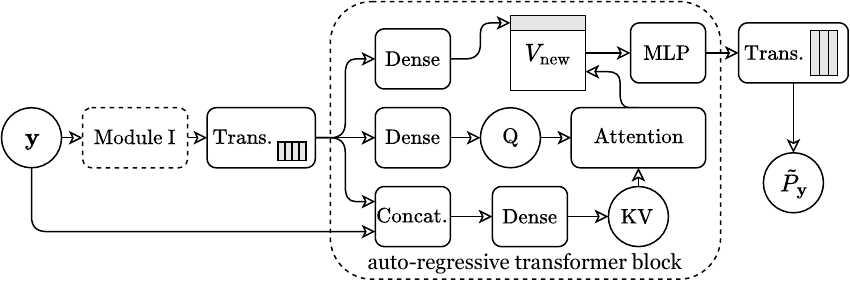}
        \label{fig:direct-access}
    }
    \caption{Architecture diagram of the different entropy models for the target representation $Y$. The direct access entropy model replaces the parameter estimator in the standalone hyper-prior entropy model with a series of transformer blocks that combine the hyper-prior and the target representation. Q, K, V corresponds to the query, key, and value embeddings in an attention mechanism. $\theta_j$ parametrizes the density function $g_w$ that generates the probability distribution assigned to the $j$-th embedding dimension of W. The Fourier basis method is the standalone hyper-prior approach with a different density function $g_w$ defining $\tilde{P}_W$. }
    \label{fig:architectures}
\end{figure*}

We construct several entropy models of varied complexity to code the intermediate representations in %auto-regressive 
transformers. The entropy models are themselves transformer-based and, although their architectures are unique, we leverage well-known concepts from existing literature~\cite{balle2018variational, MinnenBT18, LiLDLZX24} to construct them. Most learned codecs use a hyper-prior to capture high-level, global, information about the data they try to code. Although the functions generating the hyper-prior $W$ have access to the entire input data -- in our case, transformer's intermediate representation $Y$, a rate constraint placed on $W$ reduces their modeling capability. Hyper-priors are often used as side information by complex auto-regressive entropy models that also have \textit{direct access} to previously coded elements~\cite{MinnenBT18, HeYPMQW22,ZouSZ22}. This property sometimes alleviates the need for the hyper-prior.
On the other hand, early designs of learned codecs~\cite{balle2018variational} relied on entropy models that do not have direct access to previously coded elements, requiring much greater support from the \textit{standalone} hyper-prior.

We construct two entropy models with a standalone hyper-prior and one with additional direct access to the previously coded elements in the target representation. One of the entropy models with a standalone hyper-prior uses a neural network to model the data distribution (\textit{deep factorized} density model~\cite{balle2018variational}), while the other uses learned Fourier basis functions for this purpose~\cite{FuenteSB24}. The \textit{direct access} auto-regressive entropy model has access to previously coded elements of the target representation and not just the hyper-prior. Its additional complexity therefore allows for modeling more complex probability distributions. We explain the details of these models in the following subsections.

\subsection{Entropy models with standalone hyper-prior}

Entropy models of this type simplify modeling of the target representation by assuming full conditional independence of its elements given the hyper-prior. Thus, a learned entropy model:
\begin{align}
g_y : \mathcal{W} \to \mathcal{N}(\mathcal{Y} \pm \nicefrac{1}{2})
;
\mathcal{W} \subseteq \mathbb{Z}^{T \times C}
,
\mathcal{Y} \subseteq \mathbb{Z}^{T \times E}
,
\end{align}
estimates the parameters of a fully-factorized multivariate normal distribution assumed for a unit interval centered around $Y$, using $W$ as the \textit{only} context (without direct access to $Y$), effectively assuming a conditionally independent distribution for $Y$ given $W$~\cite{balle2018variational}. The rate $\smash{r_y(\mathbf{y}; \mathbf{w}): \mathcal{Y} \times\mathcal{W} \to \mathbb{R}}$ is the negative logarithm of the likelihood given by an entropy model $g_y$. See the Supplementary Material for a concrete definition.
A transformer-based hyper-prior model is given by:
\begin{align}
    W
    =
    (q \circ g_h)(Y); g_h : \mathcal{Y} \to \mathbb{R}^{T \times C}
    .
\end{align}
It is designed to enforce a dimensionality bottleneck, with each transformer block gradually decreasing the dimensions of the representation to $T \times C$. The transformer in $g_y$ gradually increases the dimensionality to $2 \times T \times E$, where the first dimension indexes the means and variances. The change of dimensionality is done by a projection layer injected between the attention layer and the multi-layer perceptron (MLP) sub-block of the transformer block. Fig.~\ref{fig:proposed} shows a generic diagram of the standalone entropy models.

The masks for the attention mechanisms in $g_h$ and $g_y$ are restricted to create representations that only depend on the current and previous time steps, such that:
\begin{align}
    \mathbf{w}_{\leq i}
    =
    g_h
    \left(
    \mathbf{y}_{\leq i}
    \right)
    \,
    \forall i \in \{1,...,T\}
    ,
\end{align}
where $\smash{\mathbf{w}_{\leq i}}$ and $\smash{\mathbf{y}_{\leq i}}$ correspond to the elements in these tensors from the first time step to the $i$-th time step.
Although this restriction is not required to code the hyper-prior or the target representation, it allows to grow the existing side information by only appending elements to it as the data is further processed. This is a critical feature for auto-regressive tasks that avoids the transmission of an entire hyper-prior for the inference of a new time frame; only the new elements generated by the new time frame need to be transmitted. Another benefit of this property is that all elements within a time frame can be coded and transmitted in parallel. This has the potential to substantially reduce the inference latency due to coding.
With this constraint, the conditional probability of the target representation given the hyper-prior is modeled as:
\begin{align}
    \textstyle
    \tilde{P}(\mathbf{y}|\mathbf{w})
    &=
    \textstyle
    \prod_{i=1}^T
    \prod_{j=1}^E
    \tilde{P}
    \left(
    y_{i,j}
    |
    \mathbf{w}_{\leq i}
    \right)
    ,
\end{align}
where $\smash{\tilde{P}}$ is a probability estimate implicitly established by the entropy model. 

The hyper-prior model $g_h$ is composed of 4 transformer blocks that sequentially bring down the embedding space $E$ to 384, 192, 96, and finally, $C = 24$ dimensions. Following~\cite{Karpathy2022}, the dense layers in the transformer blocks have no biases, and the different sub-components have residual connections and layer normalizations. The entropy model for the target representation $g_y$ is composed of another 4 transformer blocks that sequentially bring up the embedding space to 96, 192, 384, and finally, $E$ dimensions. To quantize $Y$ and $W$, $q$ is set as a conventional integer rounding operation, where, during automatic differentiation, the incoming gradients are passed to the next operation \cite{TheisSCH17}. We construct two versions of the standalone hyper-prior model: one uses a MLP to model the Cumulative Distribution Function (CDF) for each dimension of the hyper-prior, while the other uses a Fourier basis to model the corresponding Probability Density Function (PDF), as discussed next. 

\paragraph*{Deep factorized density model}
A CDF for each dimension of the hyper-prior is given by a MLP with a single sigmoid output parameterized by its weights, biases, and scaling factors of its activation functions. A monotonicity constraint is placed on these parameters to ensure the MLP produces valid CDFs and that their derivatives (the PDFs), are always non-negative~\cite{balle2018variational}.

\paragraph*{Fourier basis density model}
A PDF for each dimension of the hyper-prior is independently modeled as a Fourier series with a finite number of coefficients~\cite{FuenteSB24}. The coefficients are optimized for rate-distortion performance. To ensure non-negativity of the PDF, the coefficients are auto-correlated, making the Fourier series positive semi-definite. The function over one period is divided by its integral for normalization, and has a closed-form solution. The resulting periodic density function is extended to the entire real line by a learned mapping $(-1,1) \to \mathbb{R}$ parameterized by a scaling and an offset learned parameter. The CDF of this density function also has a simple closed-form expression. It was demonstrated in~\cite{FuenteSB24} that this model can, in some cases, better fit complex distributions compared to the deep factorized density model.

\subsection{Entropy model with direct access}
In this more complex entropy model, an auto-regressive transformer predicts the means and variances of the target representation using the hyper-prior \textit{and the previously coded elements} of the target representation. This results in the conditional probability of the target representation given the hyper-prior to be modeled as:
\begin{align}
\textstyle
\tilde{P}(\mathbf{y}|\mathbf{w})
&=
\textstyle
\prod_{i=1}^T
\prod_{j=1}^E
\tilde{P}
\left(
y_{i,j}
|
\mathbf{w}_{\leq i}
,
\mathbf{y}_{< i}
\right)
.
\end{align}
The time step restriction on $\mathbf{y}$ allows the entropy model to be used for coding, since the serialized decoding process only has access to previously-decoded elements.

Fig.~\ref{fig:direct-access} presents an overview of the architecture of the direct access entropy model. The hyper-prior $W$ is used as query embeddings and additional dimensions for the key and value embeddings of the attention mechanism in an initial transformer block. Using the same hyper-prior architecture with the deep factorized density model, the entropy model for the target representation first passes the hyper-prior through four transformer blocks with causal attention masks that maintain the same embedding size $C$. The resulting tensor is then processed by a custom transformer block in which, for each role, a single dense layer produces:
\begin{enumerate*}
    \item query embeddings of $C$ dimensions;
    \item key and value embeddings of size $C$ and $E$ respectively, as a function of the target embeddings and this tensor; and
    \item the value embedding of the first time step after attention.
\end{enumerate*}
This transformer block uses an attention mask that prevents access to elements from the current time step forward. The output of the transformer block is then further processed by three more transformer blocks that use causal attention masks and retain the same embedding size $E$, except for the last block, which increases the output embedding size to $2E$. The output tensor is split into the means and variances of the multivariate normal distribution for $Y$.

\subsection{Rate-distortion optimization}
Including the hyper-prior rate, the rate-distortion loss function is given by:
\begin{align}
    \label{eq:loss}
    \mathcal{L}
    \triangleq
    \mathbb{E}
    \left\{
    \lambda_d \, d(\hat{Z}, Z)
    +
    r_y(Y;W)
    +
    \lambda_r \, r_w(W)
    \right\}
    ,
\end{align}
where $d(\hat{Z}, Z)$ is a task distortion (loss) function, $\lambda_d \in \mathbb{R}_+$ balances the trade-off between rate (compression) and distortion (error), and $\lambda_r \in \mathbb{R}_+$ balances the trade-off between the rate of the hyper-prior and the conditional rate of the target representation. In practice we set $\lambda_r = 1$ and multiply the loss by $\lambda = \nicefrac{1}{\lambda_d}$ to obtain a single parameter in which the rate of the target representation $r_y$ and the hyper-prior $r_w$ are weighted equally.

\section{Rate-distortion performance}
\label{sec:rd-benchmarks}

\begin{figure*}[!t]
    \subfloat[Split point 6]{
        \includegraphics[width=0.485\linewidth]{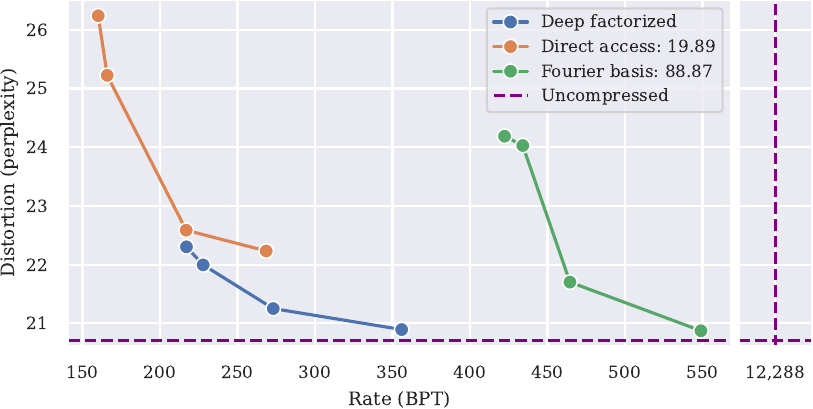}
        \label{fig:rd-hp}
    }
    \subfloat[Split points 3, 6, and 9]{
        \includegraphics[width=0.485\linewidth]{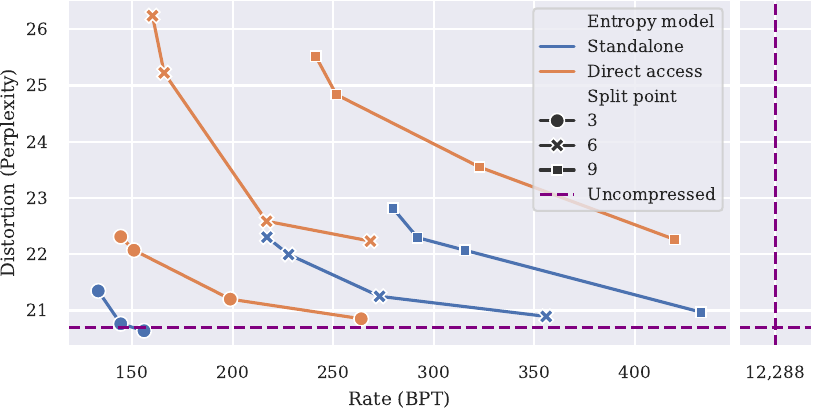}
        \label{fig:rd-sp}
    }
    \caption{Rate-distortion performance for GPT-2. The rate is measured in bits-per-token (BPT). Perplexity is the exponent of the classification cross-entropy loss, used as distortion. \textit{Uncompressed} is a model with no quantization or rate penalty ($\lambda = 0$), which uses 16 bits per token element. The entropy model with a standalone hyper-prior using the deep factorized density method outperforms the other entropy models. The rate-distortion performance of two of the entropy models decreases with the split point.\label{fig:rd}}
\end{figure*}

\begin{figure*}[!t]
    \subfloat[Split point 6]{
        \includegraphics[width=0.485\linewidth]{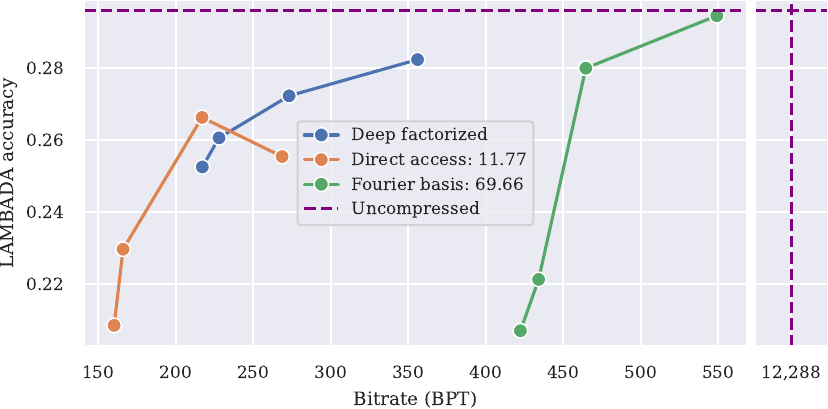}
        \label{fig:lambada-hp}
    }
    \subfloat[Split points 3, 6, and 9]{
        \includegraphics[width=0.485\linewidth]{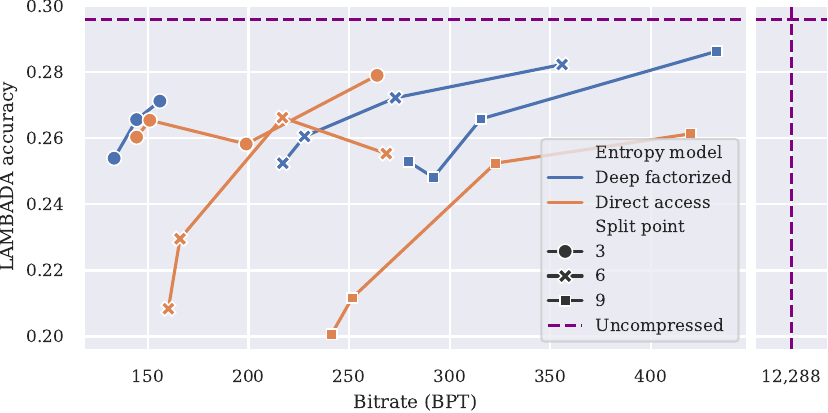}
        \label{fig:lambada-layers}
    }
    \caption{Rate-accuracy for GPT-2 evaluated on the the LAMBADA language task. The rate is measured in bits-per-token (BPT). \textit{Uncompressed} is a model with no quantization or rate penalty ($\lambda = 0$). The LAMBADA performance positively correlates with distortion (perplexity). \label{fig:lambada}}
\end{figure*}

In this section, we evaluate the rate-distortion performance on language modeling. We use the GPT-2 Small~\cite{GPT2} transformer architecture as a language model. It has $L=12$ transformer blocks with 12 attention heads and embeddings of $E=768$ dimensions.
Two language modeling tasks are considered: auto-regressive next-token prediction, for which we use the OpenWebText~\cite{Gokaslan2019OpenWeb} dataset mainly composed of Reddit conversation threads; and LAMBADA~\cite{PapernoKLPBPBBF16}, which evaluates the ability to comprehend context and understand discourse. We use the models trained on OpenWebText to evaluate their performance on LAMBADA.
Since these language models are not trained for question answering, we strip all prompts and punctuation in the LAMBADA dataset, making no distinction between context and targets.

The deep factorized entropy model has 9 dense layers with 3 hidden dimensions, for a total of 118 parameters per dimension in $C$. For the Fourier basis entropy model, we set 60 coefficients for a total of 120 parameters when counting the real and imaginary components. This is close to the number of parameters used in the deep factorized density model.

The loss function is given by Eq.~\ref{eq:loss}. We train codecs with different values of $\lambda$ at multiple split points to obtain rate-distortion curves.
All results are reported for validation sets. See the Supplementary Material for additional details. The official code is available at \href{https://github.com/adeandrade/research}{github.com/adeandrade/research}.

Fig.~\ref{fig:rd-hp} shows the rate-distortion curves for next token prediction obtained with different entropy models when the 6-th transformer block is chosen as the split point. Perplexity \cite{Perplexity} is used as the distortion measure. The rate in bits per token (BPT) is the sum of the individual rates of the hyper-prior and target representations. As seen in the figure, for equivalent perplexity, the Fourier basis and direct access entropy models require higher rate than the deep factorized model. This increase can be measured by BD-rate~\cite{bjontegaard2001calculation} and evaluates to 88.87\% and 19.89\%, respectively.
Thus, somewhat surprisingly, the simplest of the three entropy models outperforms the other two more complex models.
We provide an explanation of these findings by the theory developed in \cref{sec:rad}, which shows that the complexity of the entropy model increases the generalization error of $\mathcal{V}$-entropy and this can negatively affect the rate-distortion performance of the resulting codec.

Fig.~\ref{fig:rd-sp} shows the rate-distortion performance on next-token prediction for three equidistant split points $S$ of the standalone deep factorized hyper-prior and the direct access entropy models. Note that the deep factorized model applied to split point 3 is able to achieve the same perplexity as the transformer with uncompressed representations using only 150 bits per token, compared to 12,288 bits per token without compression. This is a 98.8\% compression ratio, on par or better than the best token pruning/compression strategies according to the recent surveys~\cite{Shao2026TMLR,Yao2026MLLMtokcomp}. Although the settings of various studies surveyed in~\cite{Shao2026TMLR,Yao2026MLLMtokcomp} are different from ours, often related to multimodal models and to reducing the total number of tokens rather than the total number of bits used up by tokens, this rough comparison shows that entropy-based strategies studied in this paper are a promising avenue for further advancement in the field of token compression. In the Supplementary Material, we present execution time analyses to supplement these results.

Another observation from Fig.~\ref{fig:rd-sp} is that, as the split point moves deeper into the transformer model, the rate-distortion performance deteriorates: more bits are required to achieve the same perplexity. This seemingly contradicts the Data Processing Inequality from information theory~\cite{Cover}, which says that the information about the input random variable cannot increase via data processing. It also seems at odds with well-known results on compression of intermediate features of CNNs~\cite{Choi2018ICIP}, where rate-distortion performance tends to improve as we move deeper into the model.
In \cref{sec:theory}, we explain this behavior by an increase in the $\mathcal{V}$-entropy of the target representation and its generalization bound. Intuitively, as the source signal is processed by non-linear functions in the model, the complexity of resulting representations can increase, negatively affecting the compression performance that a fixed set of predictive families is able to achieve.

Fig.~\ref{fig:lambada} shows LAMBADA results.
Similar to perplexity results, the deep factorized and the direct access entropy models have considerably better rate-accuracy performance than the Fourier basis model (Fig.~\ref{fig:lambada-hp}). Also, the rate-accuracy performance deteriorates the deeper we move into the transformer model (Fig.~\ref{fig:lambada-layers}). Again, this seemingly odd result comes down to the interaction between the increased complexity of representations and the limited capabilities of a predictive family upon which an entropy model is built, as we discuss in the next few sections.

\section{Rate characterization of transformers}
\label{sec:theory}

The Data Processing Inequality from information theory~\cite{Cover} states that if random variables $X$, $Y_1$, and $Y_2$ form a Markov chain $X\to Y_1 \to Y_2$, then $H(X) \geq I(X; Y_1) \geq I(X; Y_2)$, where  $I(\cdot; \cdot)$ is the mutual information. This in turn means that it should take no more bits to encode $Y_2$ than it takes to encode $Y_1$. This also holds when we fix the level of accuracy of inference from $Y_1$ or $Y_2$~\cite{Choi2022ScalableIC,HarellFADKSTAB25}. Since transformer blocks form a Markov chain, it may then seem surprising that the rate-distortion (Fig.~\ref{fig:rd-sp}) and rate-accuracy (Fig.~\ref{fig:lambada-layers}) performance deteriorates as we move deeper into the transformer.

Of course, practical (i.e., suboptimal) codecs do not necessarily behave in accordance with information-theoretic bounds. Our goal in the remainder of the paper is to provide a deeper understanding of the behavior seen in Figs.~\ref{fig:rd-sp} and~\ref{fig:lambada-layers}, since it is important for future design of efficient distributed inference systems based on transformers. We theoretically motivate two situations in which this type of behavior can be expected:
\begin{enumerate*}
    \item when the intermediate representations are expansions of the feature manifold representing the input; and
    \item when the complexity of the intermediate representations increases so that its distribution is more difficult to model.
\end{enumerate*}

The first case holds for a predictive family $\mathcal{V}$ of discretized continuous density functions, under mild assumptions. The second case holds for a wide range of possible choices for $\mathcal{V}$. This is shown in both cases using the theory of usable information. To facilitate explanations for the first case, we extend this theory by introducing the notion of a $\mathcal{V}$-entropy gap: the difference between the actual entropy of a random variable and the optimal (i.e. for an infinite sequence) rate that an entropy model for this random variable based on a predictive family $\mathcal{V}$ can achieve. This allows for explicit measurement of how suitable a predictive family can be. 

We first show that minimizing the loss function in Eq.~\ref{eq:loss} minimizes the conditional $\mathcal{V}$-entropy for the target representation $Y$, under rate and distortion constraints. We then introduce and analyze the $\mathcal{V}$-entropy gap. Using this perspective, we derive bounds for $\mathcal{V}$-entropy to show that, for a very common choice of entropy models, it is upper-bounded by the covariance of the target representation, which can increase by the expansion of the input feature space. Finally, we extend a generalization error bound for $\mathcal{V}$-entropy~\cite{XuZSSE20}, and isolate the Rademacher complexity of the target representation as part of the bound. We discuss how the generalization error affects the ability of an entropy model to learn a target distribution.

\subsection{The training objective as \texorpdfstring{$\mathcal{V}$}{V}-entropy minimization}

Since the hyper-prior $W$ is a function of the target representation $Y$, there is a trade-off between the rate of the hyper-prior and the conditional rate of the target representation, given $W$. The following theorem shows that minimizing the loss function $\mathcal{L}$ (Eq.~\ref{eq:loss}) minimizes the lower bound of the $\mathcal{V}$-entropy of the target representation $Y$ conditioned on $W$, under constraints on the rate of $W$ and the task distortion produced by $\hat{Z}$:

\begin{theorem}[store=theorem:rate]
Let $\mathcal{V}_w$ be a predictive family of entropy models $g_w$ for the hyper-prior $W$, and $\mathcal{V}_{y} \subseteq \{\mathcal{W} \cup \{ \oslash \} \to \mathcal{N}(\mathcal{Y} \pm \nicefrac{1}{2}) \}$ be a predictive family of conditional entropy models $g_y$ producing fully-factorized multivariate normal distributions discretized with a unit interval centered around the target representation $Y$. Then:
\begin{align*}
    \inf_{
    Y,
    W : H_{\mathcal{V}_w}(W) \leq R, \, \,
    \hat{Z} : \mathbb{E} \left[ d(\hat{Z}, Z) \right] \leq D
    }
    H_{\mathcal{V}_y}(Y|W)
    \leq
    \min
    \mathcal{L}
    ,
\end{align*}
where $D$ is the maximum task distortion allowed, $R$ is the maximum rate of $W$ allowed, and the minimum is over all the parameters of the functions generating the random variables in $\mathcal{L}$.
\end{theorem}
\noindent\textit{Proof.} See the Supplementary Material.

This theorem is specific to the codecs described in \cref{sec:entropy-models} because it restricts the predictive family $\mathcal{V}_y$ to discretized multivariate normal distributions with diagonal covariances, but it could be generalized to include other predictive families. The result allows us to interpret the rate-distortion optimization problem in Eq.~\ref{eq:loss} as minimizing the conditional $\mathcal{V}$-entropy under constraints. The rate restriction placed on $W$, in addition to the potential limitations of the predictive families, prevents this $\mathcal{V}$-entropy from always reaching zero. In fact, for it to be zero, all information about $Y$ must be present in $W$ in such a way that it can be coded at a rate equal to or lower (assuming $\lambda_r = 1$) than if it were to be coded in $Y$ unconditionally. Moreover, the information in $W$ must be \textit{usable} by the predictive family $\mathcal{V}_y$ so that it can reduce the entire rate of $Y$ when encoded with it.

\subsection{The \texorpdfstring{$\mathcal{V}$}{V}-entropy gap}
\label{sec:gap}

We extend the theory of \textit{usable information under computational constraints} \cite{XuZSSE20} to provide the $\mathcal{V}$-entropy gap, a measure that isolates the limitations of an entropy model, showing how much additional rate exists that is not due to the entropy of the underlying random variable.

\begin{definition}
\label{def:gap}
We define the $\mathcal{V}$-entropy gap as the difference between the rate of an infinite coding sequence achievable by the predictive family $\mathcal{V}$ and the entropy of a random variable, with both terms expressed as conditional $\mathcal{V}$-entropies with optional side information $W$:
\begin{align*}
    G_{\mathcal{V}}(Y|W)
    \triangleq
    H_{\mathcal{V}}(Y|W)
    -
    H_{\Omega}(Y|W)
    .
\end{align*}
\end{definition}
When the side information $W$ is $\oslash$, the gap can be expressed in terms of $\mathcal{V}$-entropies. Due to $\mathcal{V} \subseteq \Omega$, the $\mathcal{V}$-entropy gap is non-negative. Moreover, recall that $H_{\Omega}(Y|W) = H(Y|W)$.

The entropy gap can also be interpreted in terms of KL divergences between distributions given by the predictive family and $\Omega$. See Lemma~\ref*{lemma:kl} in the Supplementary Material for details. Hence, since we have that:
\begin{align}
    \label{eq:v-entropy-redefined}
    H_{\mathcal{V}}(Y|W)
    =
    H(Y|W)
    +
    G_{\mathcal{V}}(Y|W)
    ,
\end{align}
we can then think of $\mathcal{V}$-entropy in a similar fashion to cross-entropy, where we have the entropy of the target random variable plus a KL divergence term measuring the additional rate produced by not using the correct probability distribution.

Observing Eq.~\ref{eq:v-entropy-redefined}, it is important to note that an increase in the entropy of the target random variable $H(Y|W)$ does not necessarily increase the corresponding $\mathcal{V}$-entropy $H_{\mathcal{V}}(Y|W)$ for a fixed predictive family $\mathcal{V}$, since a more suitable entropy model could reduce the gap $G_\mathcal{V}(Y|W)$ by a larger amount than the increase in entropy. However, sufficiently robust predictive families can be expected to positively correlate with the entropy $H(Y|W)$.

\subsection{Dilation of the feature space}

In learned compression, latent representations are modeled using continuous distributions that are subsequently discretized. This fundamental choice allows the optimization of the entropy models and analysis transforms using gradient methods. In this setting, the stretch factor of the functions that generate these target representations can affect their $\mathcal{V}$-entropy regardless of the complexity of the predictive family (entropy model). We show, under mild assumptions, that the $\mathcal{V}$-entropy of a target representation $Y$ under a predictive family $\mathcal{V}$ of discretized continuous distributions is upper-bounded by the covariance determinant of the representation:

\begin{theorem}[store=theorem:covdiscrete]
Let $Y$ be a discrete random variable with sample space $\mathcal{Y}$, dimensionality $K$, means $\bm{\mu}$, and covariance matrix $\Sigma$. Let $\mathcal{V} \subseteq \Omega$ be a predictive family \cite{XuZSSE20} of probability density functions that are discretized using a fixed-step size $\Delta$. Assume that there is a probability density function $\hat{g}$ in the predictive family such that $\smash{\log \hat{g}}$ is linear or quadraric, and $\smash{\hat{g}}$ has means $\bm{\mu}$ and covariance matrix $\Sigma$ over its own support. We have:
\begin{align*}
    H_{\mathcal{V}}(Y)
    \leq
    \nicefrac{1}{2}
    \log
    \abs{\Sigma}
    (2 \pi e)^K
    -
    \log \Delta
    \quad
    \text{as}
    \;
    \Delta \to \mathbf{0}
    .
\end{align*}
\end{theorem}
\noindent\textit{Proof.} See the Supplementary Material.

Since the logarithm of the multivariate normal distribution is quadratic, the predictive family proposed in this work meets the assumption. The size of the discretization step adapts to the type of quantization performed to obtain $Y$, which in learned coding is usually done by rounding the values of $Y$ to their closest integer. With $\Delta = 1$, the provided bound is an approximation. However, this type of approximation has been used empirically with compelling results when relating Shannon's entropy and differential entropy \cite{NilssonK07, SunGWLC0S22}.
In the Supplementary Material, we present additional theoretical results and analyses that supplement Theorem~\ref{theorem:covdiscrete}.

In \cref{sec:rd-split-exp}, we estimate the covariance determinant $\abs{\Sigma}$ of the representations produced by different layers in several types of neural network: three types of transformers and a ResNet. As shown in Fig.~\ref{fig:cov}, $\abs{\Sigma}$ increases as we move deeper into transformers, but decreases as we move deeper into a ResNet. Hence, according to Theorem~\ref{theorem:covdiscrete} above, the upper bound on the $\mathcal{V}$-entropy of the latent representation increases with depth for transformers, but decreases for ResNet. This provides one reason why deeper layers in a transformer can exhibit a higher rate than shallower layers. Indeed, when $\mathcal{V}$ remains fixed throughout layers, higher rate is obtained in deeper layers in a transformer, but in the ResNet, the rate decreases with depth, as shown in Fig.~\ref{fig:layer-bpt}.

The increase (with depth) of the covariance determinant $\abs{\Sigma}$ of the representations produced by transformers is one of the main insights of this work. This increase of $\abs{\Sigma}$ is what we refer to as \textit{dilation of the feature space}: as the input signal is processed further, the feature space ``expands,'' thereby increasing the upper bound on the rate that a given predictive family $\mathcal{V}$ can achieve. This is a distinguishing feature of transformers compared to a ResNet, for example, where such dilation does not appear.

\subsection{Revisiting the generalization bound for \texorpdfstring{$\mathcal{V}$}{V}-entropy}
\label{sec:rad}

Assuming that the predictive family $\mathcal{V}$ is Lipschitz continuous, we can further upper-bound the generalization error bound (Eq.~\ref{eq:v-ent-gen-error}) in \cite{XuZSSE20} to separate the complexities of the target representation and the predictive family. Hence, we express the generalization error bound in terms of the Rademacher complexity of a set \cite{LearningTheory} for the target representation, and the Lipschitz constant of the predictive family $\mathcal{V}$:

\begin{theorem}[store=theorem:v-gen-bound]
Let $\mathcal{V} \subseteq \Omega$ be a predictive family \cite{XuZSSE20}, $Y$ and $W$ be random variables with sample spaces $\mathcal{Y}$ and $\mathcal{W}$, respectively, and $\mathcal{D} = \{(\mathbf{y}_i, \mathbf{w}_i) \}_{i=1}^N \sim Y, W$ be a set of their samples. Assume that $\forall g \in \mathcal{V}, \mathbf{y} \in \mathcal{Y}, \mathbf{w} \in \mathcal{W}, \log g[\mathbf{w}](\mathbf{y}) \in [-B, B]$, and that the functions in the predictive family are Lipschitz continuous. Then, $\forall \delta \in (0, 1)$, with probability at least $1 - \delta$, we have:
\begin{equation*}
    R_{\mathcal{V},\mathcal{D}}(Y|W)
    \leq
    2
    \,
    \mathfrak{L}(\mathcal{V}_r)
    \,
    \mathfrak{R}(\mathcal{D})
    +
    B
    \sqrt{
    \nicefrac{2}{N}
    \log \nicefrac{1}{\delta}
    }
    ,
\end{equation*}
where $\mathfrak{R}(\mathcal{D})$ is the Rademacher complexity of the concatenated samples in the dataset $\mathcal{D}$, and $\mathfrak{L}(\mathcal{V}_r)$ is the maximum Lipschitz constant in $\mathcal{V}_r = \{v| v(\mathbf{w}, \mathbf{y}) = \log g[\mathbf{w}](\mathbf{y}), g \in \mathcal{V} \}$.
\end{theorem}
\noindent\textit{Proof.} See the Supplementary Material.

In \cref{sec:rd-split-exp}, we estimate the Rademacher complexity of the target representations of various neural network layers. This quantity changes for different target representations from different layers, while the other terms present in Theorem~\ref{theorem:v-gen-bound} remain constant. We show a positive correlation between this quantity and the rate achieved for the target representations. This provides another reason -- besides feature space dilation discussed earlier -- as to why an achieved rate might increase in deeper layers of transformers. As the representation complexity increases, the generalization bound for $\mathcal{V}$-entropy worsens. This, in turn, means that higher rates might be produced by a given predictive family.  

In Corollary~\ref*{corollary:gen-bound-gaussian} in the Supplementary Material, we provide a generalization error bound for a random variable with no side information (i.e. $R_{\mathcal{V}, \mathcal{D}}(Y)$) assuming a predictive family $\mathcal{V}$ of discretized multivariate distributions with diagonal covariance matrices. We also extend the generalization error bound for $\mathcal{V}$-entropy to the $\mathcal{V}$-entropy gap, showing an additional term that accounts for the complexity of explicitly estimating the entropy of the target random variable. See Theorem~\ref*{theorem:gen-bound-gap} in the Supplementary Material. 

\section{Estimation of upper bounds}
\label{sec:rd-split-exp}

\begin{figure*}[!t]
    \subfloat[Covariance at different split points]{
        \includegraphics[width=0.485\linewidth]{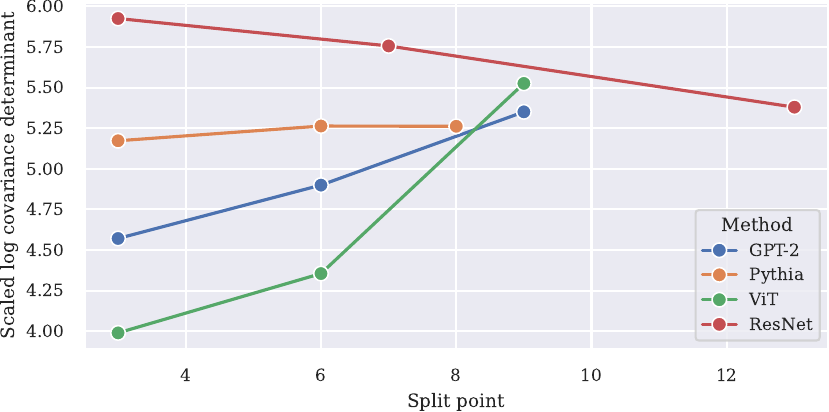}
        \label{fig:cov}
    }
    \subfloat[Covariance for corresponding rates]{
        \includegraphics[width=0.485\linewidth]{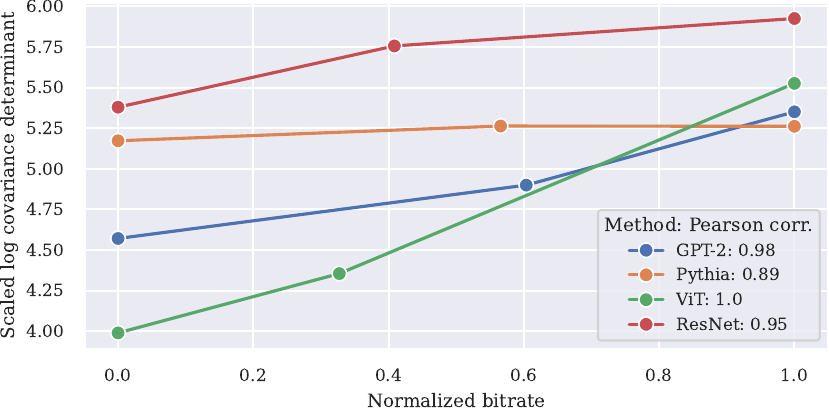}
        \label{fig:cov-bpt}
    }
    \hfil
    \subfloat[Rademacher complexity at different split points]{
        \includegraphics[width=0.485\linewidth]{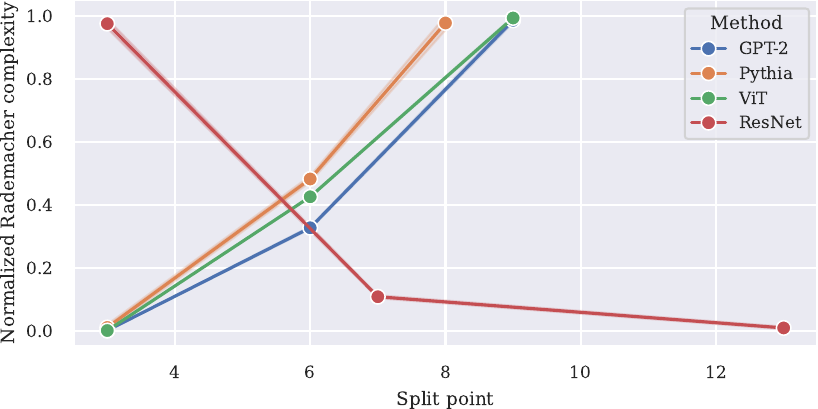}
        \label{fig:layer-rademacher}
    }
    \subfloat[Rademacher complexity for corresponding rates]{
        \includegraphics[width=0.485\linewidth]{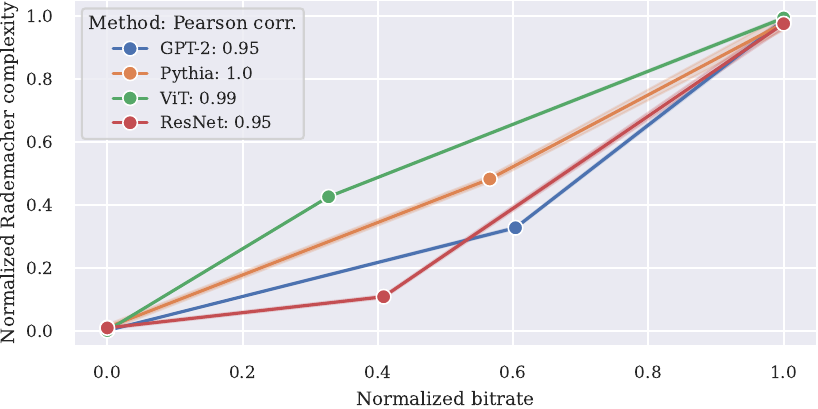}
        \label{fig:bpt-rademacher}
    }
    \caption{Rate, covariance determinant, and Rademacher complexity estimates at different split points, for GPT-2 Small, Pythia 160M, ViT B/16, and ResNet 34. Some axes are min-max scaled per method to facilitate comparison. The logarithmic scale is further scaled by $\nicefrac{1}{2D}$. The Rademacher complexity and covariance determinant strongly correlate with rate. Very narrow bands show the standard deviation of the covariance determinant and Rademacher complexity measures over 10 samples.}
\end{figure*}

The results in \cref{sec:rd-benchmarks} showed that, under similar distortion performance, the rate achieved by three different codecs on transformer's intermediaete representations increases with the split point. In \cref{sec:theory}, we explained this behavior in terms of an increase in (the upper bound on)  $\mathcal{V}$-entropy, as well as its generalization error. Since we developed upper bounds for these quantities in \cref{sec:theory}, we need to confirm this behavior empirically. Thus, in this section, we estimate the changes in the covariance determinant and Rademacher complexity of the target representation through the network. The behavior is also evaluated in other architectures and modalities, such as residual neural networks and images.

We perform experiments in which, for different tasks and model architectures, we introduce a rate constraint at different split points, measure the rate obtained by one of the entropy models from \cref{sec:entropy-models}, and compare it with estimates of the covariance determinant and the Rademacher complexity of the target representation. These two quantities are part of the $\mathcal{V}$-entropy bound in Theorem~\ref{theorem:covdiscrete} and the generalization bound in Theorem~\ref{theorem:v-gen-bound}, respectively. The entropy model architecture, settings, and the optimization algorithm, are the same across experiments, ensuring that the Lipschitz constant of the predictive family $\mathcal{V}_r$ \textit{acts as a constant} in Theorem~\ref{theorem:v-gen-bound}. In the Supplementary Material, we estimate and compare the Lipschitz constants of the entropy models obtained, showing no significant correlation with rate.

We run experiments on four types of models:
\begin{enumerate*}
    \item The GPT-2 Small language models from \cref{sec:rd-benchmarks};
    \item Language models using the Pythia 160M architecture \cite{biderman2023pythia};
    \item A transformer-based image classification task using ViT B/16 \cite{DosovitskiyB0WZ21}; and
    \item A CNN image classification task using ResNet 34 \cite{HeZRS16}.
\end{enumerate*}
The idea behind these choices is to pinpoint which aspects (i.e., model architecture, modality) produce the behavior seen in \cref{sec:rd-benchmarks} -- an increase in rate needed to achieve a certain performance -- in deeper layers of transformers. We also want to evaluate the correlation between the rate and its bounds in these diverse scenarios. A selection of key results is presented here, while more detailed results and explanations are provided in the Supplementary Material.

\subsection{Covariance determinant of the target representation}
The determinant of a covariance matrix is the product of its eigenvalues. We compute an approximation of $\abs{\Sigma}$, the covariance determinant of the target representation $Y$ (see Theorem~\ref{theorem:covdiscrete}), using the eigenvalues of the Hessenberg matrix produced by 1,000 Arnoldi iterations \cite{Stewart02a} over $N = 1,000$ samples, with context size $T = 512$. The procedure is similar to the more practical Restarted Arnoldi iteration method, popularized by the ARPACK software package \cite{lehoucq1998arpack}. The Arnoldi iteration algorithm produces an orthogonal basis for the Krylov subspace for a target matrix $A$, which is the linear subspace spanned by the images of a random vector $\mathbf{b}$ under powers of $A$ \cite{nocedal2006numerical}. The method also produces a Hessenberg (almost triangular) matrix \cite{johnson1985matrix} with the dot products of the vectors of this orthogonal basis. It is often observed that the eigenvalues of this Hessenberg matrix converge to eigenvalues of the original matrix $A$. We use this subset of eigenvalues to estimate the covariance determinant of the target representation.

\subsection{Rademacher complexity of the target representation}
To estimate the Rademacher complexity of the target representation, we replace the Rademacher random variable expectation in its definition with a sample average, obtaining:
\begin{align}
    \bar{\mathfrak{R}}(\mathcal{D})
    =
    \frac{1}{M N}
    \sum_{\mathbf{a} \in \mathcal{A}}
    \max_{i=1}^T
    \max_{j=1}^E
    \left(
    \abs{
    \sum_{k=1}^N
    a_k
    \mathcal{D}_k
    }
    \right)_{i,j},
\end{align}
where $\mathcal{D}_k$ indexes the samples of $Y$, and $\smash{\mathcal{A} = \{ \mathbf{a}_i \}_{i=1}^M \sim A}$, where $A$ is a random variable with sample space $\smash{\{-1,1\}^N}$ following the Rademacher distribution. We set $N=1,000, M=10,000$ and the context size to $T=512$. Replacing the expectation in the Rademacher complexity with its empirical estimate has been explored before in \cite{BartlettBM02}. The Rademacher complexity estimates have relatively low variance and do not significantly change the Pearson correlation coefficients.

Since our estimate of Rademacher complexity reacts to changes in dimensionality, the measure for the ResNet method is performed on the output of the convolution layer prepended to the entropy model, which has the same dimensionality across split points. The target representation is quantized before computing the Rademacher complexity and covariance determinant estimates. 

\subsection{Results}

\begin{figure}[!t]
    \includegraphics[width=\linewidth]{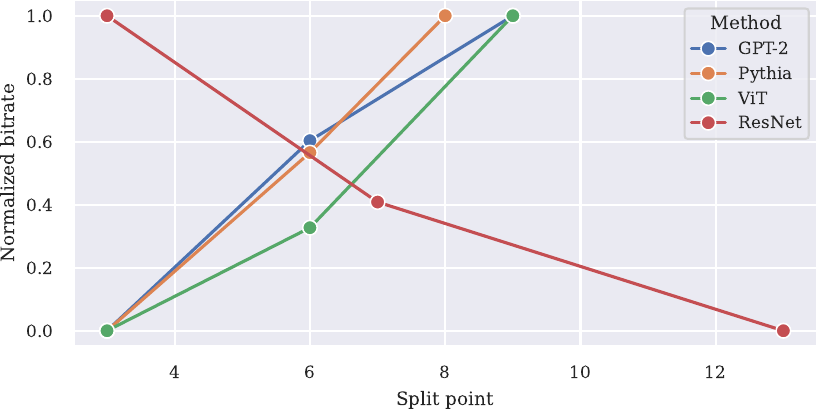}
    \caption{Rate at different split points. Only in ResNets, the achieved rate decreases with the split point. The bitrate is min-max scaled per method to facilitate comparison on the same graph.}
    \label{fig:layer-bpt}
\end{figure}

Fig.~\ref{fig:cov} shows estimates of the covariance determinant of the target representation at different split points, while Fig.~\ref{fig:layer-bpt} shows the bitrates\footnote{Normalized to the highest bitrate, for easier plotting.} obtained by the standalone deep factorized model at the various split points. As seen in these figures, the achieved bitrate increases with the split point in transformers, but not in the ResNet.  Similarly, the covariance determinant of the target representation increases with the split point in transformers, but not in the ResNet. When the covariance determinant is plotted against the normalized bitrate (Fig.~\ref{fig:cov-bpt}), the positive correlation trend is evident. In fact, the average Pearson correlation between these two measurements is 0.96. This strong correlation corroborates the relationship between the $\mathcal{V}$-entropy (i.e., the achieved rate) and its upper bound provided in Theorem~\ref{theorem:covdiscrete}. It also provides an explanation for the phenomenon of rate increase in deeper layers of transformers. 

Fig.~\ref{fig:layer-rademacher} shows that the Rademacher complexity of the intermediate representations increases with depth in transformers, but not in ResNets. Finally, Fig.~\ref{fig:bpt-rademacher} plots the estimate of the Rademacher complexity against the normalized bitrate of the target representation. Again, we see a strong positive correlation trend, with an average Pearson correlation of 0.97. These results provide empirical illustration for Theorem~\ref{theorem:v-gen-bound}. Specifically, as the Rademacher complexity of the target representation increases, so does the generalization bound for the rate achieved by an entropy model trained on that representation.

\section{Summary and conclusion}

In this paper, we studied the rate-distortion performance of transformers in the context of distributed inference. In particular, the intermediate representations of transformers were compressed, and its effect on task performance was analyzed. We presented several lossy compression methods for this purpose, with particular emphasis on entropy modeling, an aspect that has received limited attention in token pruning (quantization) research thus far. These methods achieved excellent compression results -- close to 99\% compression with no loss in perplexity -- which is on par or better than the best token pruning (quantization) approaches available. 

Moreover, our experiments revealed an interesting phenomenon: rate-distortion performance degrades as we compress deeper and deeper representations in transformers. This is contrary to what has been observed in CNNs, and also seemingly violates well-known results from information theory. To explain this phenomenon, we drew concepts from the theory of usable information, statistical learning, as well as classical rate-distortion theory. We showed that rate-distortion optimization can be considered as $\mathcal{V}$-entropy minimization, making $\mathcal{V}$-entropy a central concept in understanding rate-distortion behavior of learned codecs. We then derived upper bounds on $\mathcal{V}$-entropy and its generalization error, in terms of the covariance determinant and the Rademacher complexity of the target representation. We also demonstrated experimentally that the covariance determinant and the Rademacher complexity of the target representation increase with depth in transformers, but not in a ResNet. Together, these results explain the observed rate-distortion behavior in transformers, and why it is different from the behavior observed on CNNs. 

Our experiments using methods that constrain the Lipschitz constant of the function generating the target representation -- a quantity that bounds both its covariance determinant and Rademacher complexity -- result in language models for which training does not effectively reduce the task loss. This corroborates the results of \cite{Newhouse2025TrainingTW}, where, to match the validation accuracy of NanoGPT \cite{Karpathy2022}, the constant of a Lipschitz-constrained transformer had to be relaxed to be very high.

On the practical side, our resuts suggest that shallower representations (earlier split points) in transformers are easier to compress with codecs of limited complexity. This is advantageous in the IoT domain, where resource-constrained edge devices are limited in both the number of inference steps they can perform, and the complexity of codecs they can use. Conveniently then, transmitting representations from an early split point would result in better rate-distortion performance while also accommodating the resource constraints. Powerful devices on the cloud would then perform the rest of the inference process.

A commensurate increase in the complexity of the entropy model to meet the requirements of a target representation could result in a decrease in its $\mathcal{V}$-entropy. However, this positive contribution is offset by an increase in its generalization error. This might explain why the more complex entropy models studied in this work do not perform as well as the simplest entropy model. This result reinforces the need for codecs with better inductive biases that exploit properties of the data to increase performance while remaining simple. In particular, alternatives to current learnable codecs that do not assume a continuous distribution over the target representation could overcome increases in $\mathcal{V}$-entropy in transformer's deeper representations due to the potential invariance of these codecs to dilations of the feature space.

% argument is your BibTeX string definitions and bibliography database(s)
\bibliographystyle{IEEEtran}
\bibliography{references}

\newpage

\begin{IEEEbiography}[{\includegraphics[width=1in,height=1.25in,clip,keepaspectratio]{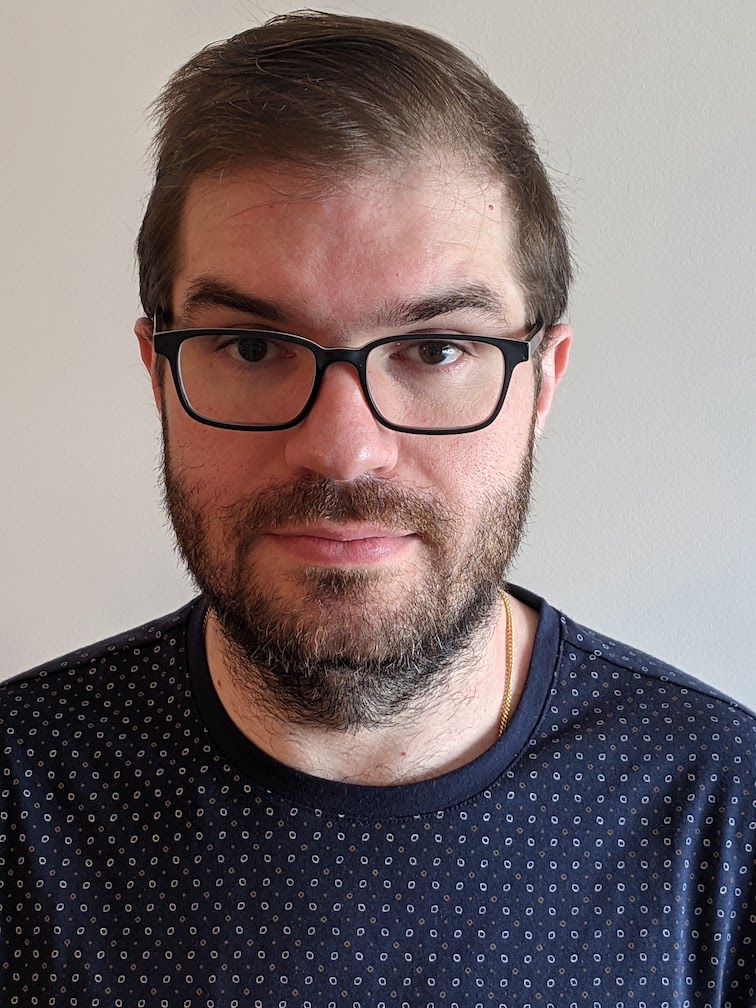}}]{Anderson de Andrade} (S'22) received his M.Sc. in Applied Computing from the University of Toronto in 2015 and obtained a B.Eng. degree in Networks and Communications in 2007 from Universidad Tecnológica del Centro. He is currently an Engineering Science Ph.D. student at Simon Fraser University. His research interests include learned compression, information theory, and learning theory. He has published at major conferences, including ICLR, and EMNLP, and has been awarded the NSERC CGS-D scholarship.
\end{IEEEbiography}

\begin{IEEEbiography}[{\includegraphics[width=1in,height=1.25in,clip,keepaspectratio]{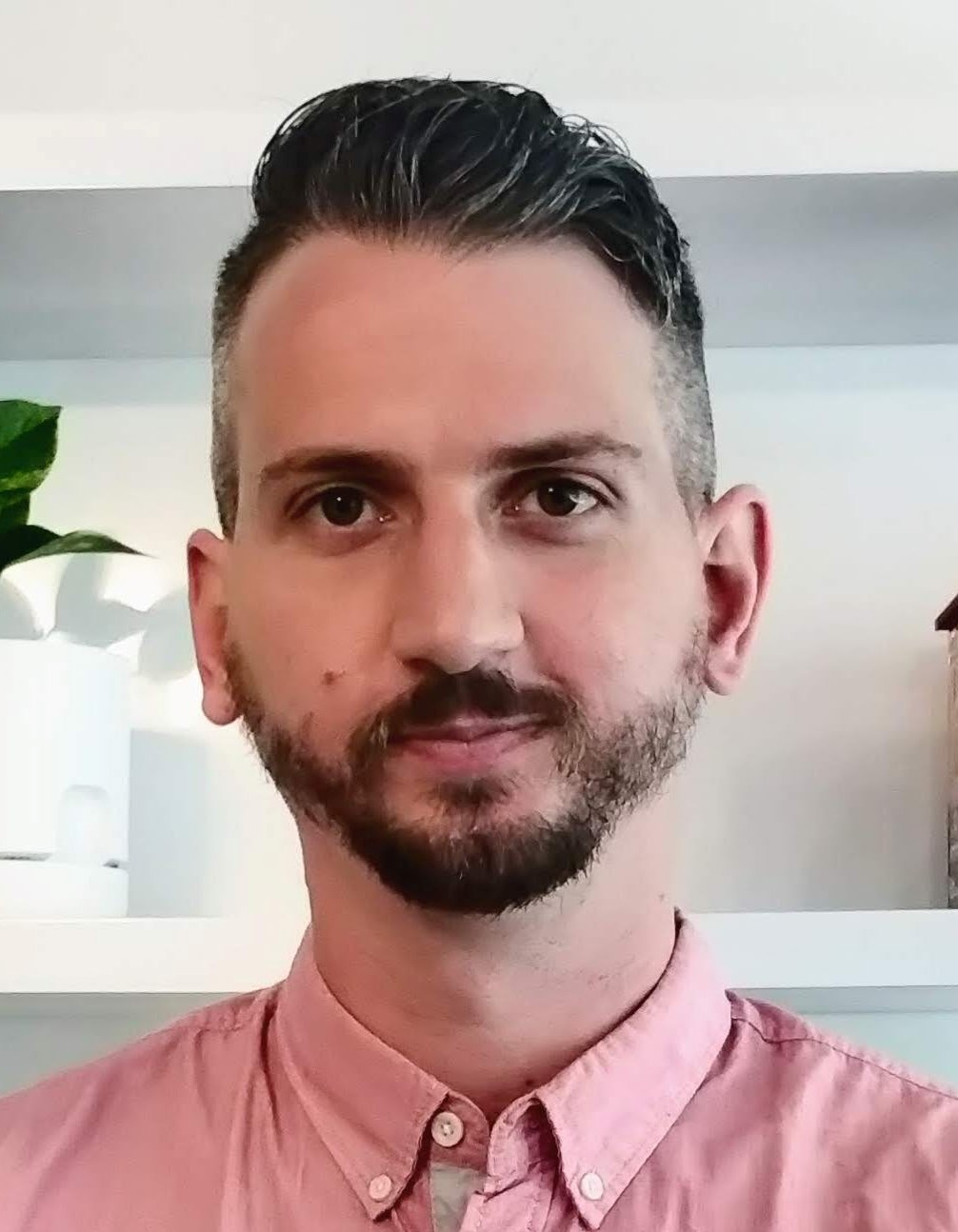}}]{Alon Harell}(S'19)
received the M.A.Sc. degree in electrical engineering from Simon Fraser University, Burnaby, BC, Canada  in 2020 focusing on deep learning applications for non-intrusive load monitoring. 
Since 2020 Alon has been pursuing his PhD in engineering science at Simon Fraser University. His research interests include information theory as it applies to deep learning, coding for machines, and sports analytics. He has published at major conferences including ICASSP, ICM Multimedia, and AAAI, and has been awarded both NSERC CGS-M and PGS-D scholarships.
\end{IEEEbiography}

\begin{IEEEbiography}[{\includegraphics[width=1in,height=1.25in,clip,keepaspectratio]{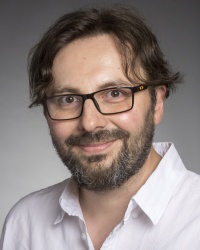}}]{Ivan V. Baji\'{c}}(S'99--M’04--SM’11) is a Professor of Engineering Science and co-director of the Multimedia Lab at Simon Fraser University, Canada. His research interests include signal processing and machine learning with applications to multimedia signal processing, compression, and collaborative intelligence. His group’s work has received the 2023 IEEE TCSVT Best Paper Award, conference paper awards at ICME 2012, ICIP 2019,  MMSP 2022, and ISCAS 2023, and other recognitions (e.g., paper award finalist, top n\%) at Asilomar, ICIP, ICME, and CVPR. He is the Past Chair of the IEEE Multimedia Signal Processing Technical Committee and currently serves as a Senior Area Editor of IEEE Signal Processing Letters.
\end{IEEEbiography}

\onecolumn

\setcounter{section}{0}
\setcounter{theorem}{3}
\setcounter{definition}{1}
\setcounter{equation}{12}

\begin{center}
\Large{\textbf{Supplementary Material}}
\end{center}

\etocdepthtag.toc{appendix}
\etocsettagdepth{main}{none}
\etocsettagdepth{appendix}{subsubsection}
\tableofcontents

\section{Mathematical notation}

\begin{table}[!t]
\centering
\caption{Notation reference}
\label{table:notation}
\begin{tabular}{ll}
    \toprule
    Notation & Definition \\
    \midrule
    $\mathbf{a} \sim A$ & Rademacher random variable with sample space $\smash{\{-1,1\}^N}$ \\
    $\mathcal{A}$ & $M$ samples of $A$: $\smash{\mathcal{A} = \{ \mathbf{a}_i \}_{i=1}^M \sim A}$
    \\
    $B$ & Bound such that $\log g[\mathbf{w}](\mathbf{y}), \log P_{Y|W}(\mathbf{y}|\mathbf{w}) \in [-B, B]$ \\
    $C$ & Embedding size of the side information: $W \in \mathbb{R}^{T \times C}$ \\
    $D$ & Distortion level \\
    $\mathcal{D}$ & $N$ data samples: $\mathcal{D} = \{(\mathbf{y}_i, \mathbf{w}_i) \}_{i=1}^N \sim Y, W$ \\
    $d(\mathbf{\hat{z}}, \mathbf{z})$ & Task loss function or distortion function \\
    $E$ & Embedding size of the target representation: $Y \in \mathbb{R}^{T \times E}$ \\
    $f_{1,S}(X)$ & Function producing the target representation $Y$ \\
    $\smash{f_{S+1,L}(Y)}$ & Second split/module of the neural network, without header layers \\
    $f_l(\cdot)$ & Transformer block: $\{ f_l : \mathbb{R}^{T \times E} \to \mathbb{R}^{T \times E} \}_{l=1}^{L}$ \\
    $G_{\mathcal{V}}(\cdot|\cdot)$ & The $\mathcal{V}$-entropy gap, Definition~\ref*{def:gap} \\
    $\smash{g_\mu(W)}$ & Means produced by $g_y$, $\smash{g_\mu : \mathcal{W} \to \mathbb{R}^{T \times E}}$ \\
    $\smash{g_\sigma(W)}$ & Variances produced by $g_y$, $\smash{g_\sigma : \mathcal{W} \to \mathbb{R}_{+}^{T \times E}}$ \\
    $g_h(Y)$ & Hyper-prior model $\smash{g_h : \mathcal{Y} \to \mathbb{R}^{T \times C}}$ \\
    $g_t(\cdot)$ & $\smash{g_t : \mathbb{R}^{T \times E} \to \mathcal{Z}}$, head module producing predictions $\smash{\hat{Z}}$ for target $Z$ \\
    $\smash{g_y(W)}$ & Entropy model for the target representation $\smash{g_y : \mathcal{W} \to \mathcal{N}(\mathcal{Y})}$ 
    \\
    $\smash{g_w(w_{i,j}; \bm{\theta}_j)}$ & Entropy model for the hyper-prior $g_w : \mathbb{R} \to [0, 1]$ \\
    $H_{\mathcal{V}}(\cdot | \cdot)$ & Conditional $\mathcal{V}$-entropy, Eq.~\ref*{eq:conditional-entropy} \\
    $H(\cdot)$ & Shannon's entropy \\
    $H(\cdot, \cdot)$ & Cross-entropy \\
    $h(\cdot)$ & Differential entropy \\
    $h(\cdot, \cdot)$ & Differential cross-entropy \\
    $I$ & Identity matrix \\
    $J_f(\mathbf{y})$ & Jacobian matrix of function $f$ evaluated at $\mathbf{y}$ \\
    $K$ & Number of elements in $Y$ such that $K = T \times E$ \\
    $\mathrm{KL}(\cdot \Vert \cdot)$ & Kullback–Leibler divergence \\
    $L$ & Number of transformer blocks in a transformer-based neural network \\
    $\smash{\mathcal{L}}$ & Loss function, Eq.~\ref*{eq:loss} \\
    $M$ & Number of samples from $A$: $\abs{\mathcal{A}}$ \\
    $\mathcal{N}, \mathcal{N}(\mathcal{Y})$ & Normal PDF, or the set of all normal PDFs on $\mathcal{Y}$ \\
    $N$ & Dataset size $\abs{\mathcal{D}}$ \\
    $q(\cdot)$ & Quantization function \\
    $R$ & Rate value \\
    $R_{\mathcal{V},\mathcal{D}}(\cdot|\cdot)$ & Generalization error of $\mathcal{V}$-entropy, Eq.~\ref*{eq:v-ent-gen-error} \\
    $S_{\mathcal{V},\mathcal{D}}(\cdot|\cdot)$ & Generalization error of the $\mathcal{V}$-entropy gap, Definition~\ref{def:gen} \\
    $r_y(\mathbf{y}; \mathbf{w})$ & Rate function for the target representation, Eq.~\ref{eq:rate} \\
    $r_w(\mathbf{w})$ & Rate function for the side-information, Eq.~\ref{eq:hp-rate} \\
    $S$ & Split point $S \in \{1, ..., L \}$\\
    $T$ & Target representation context size: $Y \in \mathbb{R}^{T \times E}$ \\
    $\mathcal{V}$ & Predictive family \cite{XuZSSE20} \\
    $\mathcal{V}_r$ & Set of log probability functions of $\mathcal{V}$: $\{v| v(\mathbf{w}, \mathbf{y}) = \log g[\mathbf{w}](\mathbf{y}), g \in \mathcal{V} \}$ \\
    $\mathbf{w} \sim W, \mathbf{w} \in \mathcal{W}$ & Side information with sample space $\smash{\mathcal{W} \subseteq \mathbb{R}^{T \times C}}$ \\
    $\mathbf{x} \sim X, \mathbf{x} \in \mathcal{X}$ & Input with $\smash{\mathcal{X} \subseteq \mathbb{R}^{T \times E}}$ \\
    $\mathbf{y} \sim Y, \mathbf{y} \in \mathcal{Y}$ & Target representation with $\smash{\mathcal{Y} \subseteq \mathbb{R}^{T \times C}}$ \\
    $\mathbf{z} \sim Z, \mathbf{z} \in \mathcal{Z}$ & Task target \\
    $\smash{\mathbf{\hat{z}} \sim \hat{Z}}, \mathbf{\hat{z}} \in \mathcal{Z}$ & Model prediction \\
    $\Delta$ & Quantization step \\
    $\Theta$ & Parameters of the hyper-prior entropy model: $\Theta = \{ \bm{\theta}_j, ..., \bm{\theta}_C \}$ \\
    $\lambda$ & Rate-distortion trade-off parameter \\
    $\bm{\mu}$ & Means of $Y$ in Theorem~\ref*{theorem:covdiscrete} \\
    $\Sigma$ & Covariance of $Y$ in Theorem~\ref*{theorem:covdiscrete} \\
    $\Phi$ & Normal CDF \\
    $\Omega$ & Set of all probability functions over $\mathcal{Y}$
    such that $\Omega = \{\mathcal{W} \cup \{ \oslash \} \to \mathcal{P}(\mathcal{Y}) \}$ \\
    \bottomrule
\end{tabular}
\end{table}

\noindent Table~\ref{table:notation} compiles the most relevant mathematical notation used in this work.

\section{Theoretical results and proofs}

\subsection{Training objective as \texorpdfstring{$\mathcal{V}$}{V}-entropy minimization}

\getkeytheorem{theorem:rate}
\begin{proof}
Establishing that the infimum is over $
Y,
W : H_{\mathcal{V}_w}(W) \leq R,
\hat{Z} : \mathbb{E} [ d(\hat{Z}, Z) ] \leq D
$,
we have:
\begin{align}
    \inf
    H_{\mathcal{V}_y}(Y|W)
    \label{eq:lagrangian-relaxation}
    &\leq
    \min_{Y, W, \hat{Z}}
    \left\{
    \lambda_d
    \left(
    \mathbb{E} \left[ d(\hat{Z}, Z) \right]
    -
    D
    \right)
    +
    H_{\mathcal{V}_y}(Y|W)
    +
    \lambda_r
    \left(
    H_{\mathcal{V}_w}(W)
    -
    R
    \right)
    \right\}
    \\
    &\leq
    \min_{Y, W, \hat{Z}}
    \left\{
    \lambda_d
    \,
    \mathbb{E} \left[ d(\hat{Z}, Z) \right]
    +
    \inf_{g_y \in \mathcal{V}_y}
    \mathbb{E}
    \left[
    - \log
    g_y[W]
    \left(
    Y
    \right)
    \right]
    +
    \lambda_r
    \,
    H_{\mathcal{V}_w}(W)
    \right\}
    \\
    &\leq
    \min_{Y, W, \hat{Z}, g_y \in \mathcal{V}_y}
    \left\{
    \lambda_d
    \,
    \mathbb{E} \left[ d(\hat{Z}, Z) \right]
    +
    \mathbb{E}
    \left[
    - \log
    g_y[W]
    \left(
    Y
    \right)
    \right]
    +
    \lambda_r
    \inf_{g_w \in \mathcal{V}_w}
    \mathbb{E}
    \left[
    - \log g_w(W)
    \right]
    \right\}
    \\
    &\leq
    \min_{Y, W, \hat{Z}, g_y \in \mathcal{V}_y, g_w \in \mathcal{V}_w}
    \left\{
    \lambda_d
    \,
    \mathbb{E} \left[ d(\hat{Z}, Z) \right]
    +
    \mathbb{E}
    \left[
    - \log
    g_y[W]
    \left(
    Y
    \right)
    \right]
    +
    \lambda_r
    \,
    \mathbb{E}
    \left[
    - \log g_w(W)
    \right]
    \right\}
    \\
    &=
    \min
    \mathcal{L}
    .
\end{align}
Eq.~\ref{eq:lagrangian-relaxation} uses the Lagrangian relaxation solution as a bound \cite{HiriartUrruty1993ConvexAA}. $D$ and $R$ are dropped since they are non-negative.
\end{proof}

For a fixed random variable $Y$, we can think of the target objective of Theorem~\ref*{theorem:rate} as the minimization of the $\mathcal{V}$-entropy gap under similar constraints, since the entropy of $Y$ is not being optimized. If the analysis transform that produces $Y$ was lossless (e.g. invertible), this would be the case as well.

\subsection{The \texorpdfstring{$\mathcal{V}$}{V}-entropy gap in terms of KL divergences}

\begin{lemma}
\label{lemma:kl}
Let $\mathcal{V} \subseteq \Omega$ be a predictive family \cite{XuZSSE20}, and $Y$ and $W$ be two random variables with sample space $\mathcal{Y}$ and $\mathcal{W}$ respectively. Then: 
\begin{align}
    G_{\mathcal{V}}(Y|W)
    =
    \inf_{g \in \mathcal{V}}
    \mathbb{E}_{\mathbf{w} \sim W}
    \left[
    \mathrm{KL}
    \left(
    P_{Y|\mathbf{w}}
    \Vert
    g[\mathbf{w}]
    \right)
    \right]
    .
\end{align}

\begin{proof}
\begin{align}
    G_{\mathcal{V}}(Y|W)
    &
    \triangleq
    H_{\mathcal{V}}(Y|W)
    -
    H_{\Omega}(Y|W)
    \\
    &=
    \inf_{g \in \mathcal{V}}
    \mathbb{E}_{\mathbf{y},\mathbf{w} \sim Y,W}
    \left[
    -
    \log
    g[\mathbf{w}](\mathbf{y})
    \right]
    -
    \inf_{\omega \in \Omega}
    \mathbb{E}_{\mathbf{y},\mathbf{w} \sim Y,W}
    \left[
    -
    \log
    \omega[\mathbf{w}](\mathbf{y})
    \right]
    \\
    &=
    \label{eq:inf}
    \inf_{g \in \mathcal{V}}
    \mathbb{E}_{\mathbf{y},\mathbf{w} \sim Y,W}
    \left[
    -
    \log
    g[\mathbf{w}](\mathbf{y})
    \right]
    -
    \mathbb{E}_{\mathbf{y},\mathbf{w} \sim Y,W}
    \left[
    -
    \log
    P_{Y|W}(\mathbf{y}|\mathbf{w})
    \right]
    \\
    &=
    \inf_{g \in \mathcal{V}}
    \mathbb{E}_{\mathbf{w} \sim W}
    \left[
    \mathbb{E}_{\mathbf{y} \sim Y|\mathbf{w}}
    \left[
    \log
    \frac{
    P_{Y|W}(\mathbf{y}|\mathbf{w})
    }{
    g[\mathbf{w}](\mathbf{y})
    }
    \right]
    \right]
    \\
    &=
    \inf_{g \in \mathcal{V}}
    \mathbb{E}_{\mathbf{w} \sim W}
    \left[
    \mathrm{KL}
    \left(
    P_{Y|\mathbf{w}}
    \Vert
    g[\mathbf{w}]
    \right)
    \right]
    .
\end{align}
Eq.~\ref{eq:inf} was obtained due to $
\forall \mathbf{w} \in \mathcal{W}
,
P_{Y|\mathbf{w}} \in \Omega
$
and
$
H(P_{Y|\mathbf{w}},P_{Y|\mathbf{w}}) 
\leq
H(P_{Y|\mathbf{w}},\omega[\mathbf{w}])$, where $H(\cdot, \cdot)$ is the cross-entropy.
\end{proof}
\end{lemma}

\subsection{Dilation of the feature space}

\subsubsection{\texorpdfstring{$\mathcal{V}$}{V}-entropy bound for a predictive family of discretized continuous distributions}

We present one of the main results in this work, as discussed in the paper.

\getkeytheorem{theorem:covdiscrete}
\begin{proof}
Let $\mathcal{\bar{V}}$ be a portion of the predictive family $\mathcal{V}$ that skips the discretization step. Note that we have $\hat{g} \in \mathcal{\bar{V}}$. Let $f_Y$ be a generalized probability density function for $P_Y$ using a Dirac delta representation:
\begin{align}
    \label{eq:dirac}
    f_Y(\mathbf{y})
    =
    \textstyle
    \sum_{\mathbf{\bar{y}} \in \mathcal{Y}}
    P_Y(\mathbf{\bar{y}})
    \delta(\mathbf{y} - \mathbf{\bar{y}})
    .
\end{align}
We have the following:
\begin{align}
    H_{\mathcal{V}}(Y)
    &\triangleq
    \inf_{g \in \mathcal{V}}
    -
    \sum_{\mathbf{y} \in \mathcal{Y}}
    P_Y(\mathbf{y})
    \log
    g[\oslash](\mathbf{y})
    \\
    &=
    \inf_{\bar{g} \in \mathcal{\bar{V}}}
    -
    \sum_{\mathbf{y} \in \mathcal{Y}}
    P_Y(\mathbf{y})
    \log
    \int_{\mathbf{y} - \nicefrac{\Delta}{2}}^{\mathbf{y} + \nicefrac{\Delta}{2}}
    \bar{g}[\oslash](\mathbf{u})
    \,
    du
    \\
    \label{eq:mvt}
    &=
    \inf_{\bar{g} \in \mathcal{\bar{V}}}
    -
    \sum_{\mathbf{y} \in \mathcal{Y}}
    P_Y(\mathbf{y})
    \log
    \left[
    \bar{g}[\oslash](\mathbf{y})
    \Delta
    \right]
    \quad
    \text{as}
    \;
    \Delta \to \mathbf{0}
    \\
    &=
    \inf_{\bar{g} \in \mathcal{\bar{V}}}
    -
    \sum_{\mathbf{y} \in \mathcal{Y}}
    P_Y(\mathbf{y})
    \log
    \bar{g}[\oslash](\mathbf{y})
    -
    \log \Delta
    \quad
    \text{as}
    \;
    \Delta \to \mathbf{0}
    \\
    \label{eq:dirac-use}
    &=
    \inf_{\bar{g} \in \mathcal{\bar{V}}}
    -
    \int_{\mathbb{R}^K}
    f_Y(\mathbf{y})
    \log
    \bar{g}[\oslash](\mathbf{y})
    \,
    d\mathbf{y}
    -
    \log \Delta
    \quad
    \text{as}
    \;
    \Delta \to \mathbf{0}
    \\
    &\leq
    -
    \int_{\mathbb{R}^K}
    f_Y(\mathbf{y})
    \log
    \hat{g}[\oslash](\mathbf{y})
    \,
    d\mathbf{y}
    -
    \log \Delta
    \quad
    \text{as}
    \;
    \Delta \to \mathbf{0}
    \\
    &=
    -
    \mathbb{E}_{\mathbf{y} \sim f_Y}
    \left[
    c_1 \mathbf{y}^2
    +
    c_2 \mathbf{y}
    +
    c_3
    \right]
    -
    \log \Delta
    \quad
    \text{as}
    \;
    \Delta \to \mathbf{0}
    \\
    &=
    -
    \left[
    c_1
    (
    \Sigma
    +
    \bm\mu \bm\mu^\top
    )
    +
    c_2 \bm\mu
    +
    c_3
    \right]
    -
    \log \Delta
    \quad
    \text{as}
    \;
    \Delta \to \mathbf{0}
    \\
    \label{eq:moments-match}
    &=
    -
    \mathbb{E}_{\mathbf{y} \sim \hat{g}}
    \left[
    c_1 \mathbf{y}^2
    +
    c_2 \mathbf{y}
    +
    c_3
    \right]
    -
    \log \Delta
    \quad
    \text{as}
    \;
    \Delta \to \mathbf{0}
    \\
    &=
    -
    \int_{\mathbb{R}^K}
    \hat{g}[\oslash](\mathbf{y})
    \log
    \hat{g}[\oslash](\mathbf{y})
    \,
    d\mathbf{y}
    -
    \log \Delta
    \quad
    \text{as}
    \;
    \Delta \to \mathbf{0}
    \\
    &=
    h[\bar{g}]
    -
    \log \Delta
    \quad
    \text{as}
    \;
    \Delta \to \mathbf{0}
    \\
    \label{eq:norm-upper-bound}
    &\leq
    h
    \left[
    \mathcal{N}
    \left(
    \bm{\mu},\Sigma
    \right)
    \right]
    -
    \log \Delta
    \quad
    \text{as}
    \;
    \Delta \to \mathbf{0}
    \\
    &=
    \label{eq:entropy-norm}
    \nicefrac{1}{2}
    \log
    \abs{\Sigma}
    (2 \pi e)^K
    -
    \log \Delta
    \quad
    \text{as}
    \;
    \Delta \to \mathbf{0}
    ,
\end{align}
where $c_1$, $c_2$, and $c_3$ are the coefficients of the polynomial $\log \hat{g}$ of, at most, degree 2, and $h[\cdot]$ is the differential entropy of a probability density function (PDF) over its support.

Eq.~\ref{eq:mvt} uses the mean value theorem and shrinks the step size to obtain $\bar{g}[\oslash](\mathbf{y})$ as the probability of the entropy model given to the symbol $\mathbf{y}$. Eq.~\ref{eq:dirac-use} uses the generalized probability density function (Eq.~\ref{eq:dirac}) to change the entropy measure to differential entropy. Eq.~\ref{eq:moments-match} uses the fact that the means and covariances of $Y$ and $\hat{g}$ match. Eq.~\ref{eq:norm-upper-bound} uses the differential entropy of a Gaussian distribution with means $\bm\mu$ and covariances $\Sigma$ as an upper-bound on the differential entropy of any random variable with the same means and covariances \cite{Cover}. Eq.~\ref{eq:entropy-norm} uses the definition of differential entropy for multivariate normal distributions.
\end{proof}

Supplementing Theorem~\ref*{theorem:covdiscrete}, we provide lower-bounds for differential entropy under bijectivity assumptions of the functions that produce the target representation and the side information. The bounds show interesting relationships between the Lipschitz constants of these functions and the differential entropy of the target representation. Since the differential entropy lower-bounds the covariance of the same random variable \cite{Cover}, by Theorem~\ref*{theorem:covdiscrete}, these bounds can positively correlate with $\mathcal{V}$-entropy. If we have side information that originates from a common ancestor with the target random variable, we have the bound shown by Theorem~\ref{theorem:ca}. If the side information $W$ is a function of the target representation $Y$, as is the case in the proposed entropy models, we have the bound of Theorem~\ref{theorem:fixed}.

\subsubsection{Entropy bound for a discretized transformed continuous random variable}

In situations where the input is \textit{truly} (i.e. not assumed) continuous, the quantization of the transformed (target) representation is the discretization of this signal. The entropy of this quantized random variable correlates with the differential entropy of the pre-quantized continuous variable. Moreover, the differential entropy of this target representation is also affected by any expansion caused by the function that produces it, since the differential entropy of a continuous variable is not invariant under change of variables \cite{Cover}. We show that the Lipschitz constant -- a direct measure of the dilation of a function \cite{benyaminigeometric} -- of the function generating a target representation can increase its entropy. More generally, we show that an expansion of a continuous feature space increases the entropy of the resulting discretized random variable:

\begin{theorem}
Let $\smash{Y = q(Y'; \Delta)}$ be the quantization of a $L_f$-Lipschitz function of a continuous random variable $X$ such that $Y' = f(X)$, where $\Delta$ is the size of a fixed quantization step, and $K$ is the dimensionality of $X$ and $Y$. Then, we have:
\begin{equation*}
    H(Y)
    \leq
    h(X)
    +
    (L_f - 1)K
    -
    \log \Delta
    \quad
    \text{as}
    \;
    \Delta \to 0
    ,
\end{equation*}
where $h(X)$ is the differential entropy of $X$.

\begin{proof}
Starting from Eq.~\ref{eq:diff-vs-ent}, we have:
\begin{align}
    H(Y)
    \label{eq:diff-vs-ent}
    &\leq
    h(Y')
    -
    \log \Delta
    \quad
    \text{as}
    \;
    \Delta \to 0
    \\
    \label{eq:diff-ent-jac}
    &\leq
    h(X)
    +
    \mathbb{E}
    \left[
    \log
    \abs{
    J_{f(X)}
    }
    \right]
    -
    \log \Delta
    \quad
    \text{as}
    \;
    \Delta \to 0
    \\
    \label{eq:diff-lips}
    &\leq
    h(X)
    +
    (L_f - 1)K
    -
    \log \Delta
    \quad
    \text{as}
    \;
    \Delta \to 0
    .
\end{align}
Eq.~\ref{eq:diff-vs-ent} uses $H(Y) \to h(Y') - \log \Delta$ as $\Delta \to 0$ \cite{Cover}. Eq.~\ref{eq:diff-ent-jac} uses an well-known upper-bound on the entropy of a function of a continuous random variable \cite{DiffEntropy}. Eq.~\ref{eq:diff-lips} uses the same derivation that produces Eq.~\ref{eq:lips-from-jac} from Eq.~\ref{eq:log-jac}.
\end{proof}
\end{theorem}

As discussed in \cref*{sec:gap}, an increase in the entropy of the target representation could increase its $\mathcal{V}$-entropy. This result shows that this can occur even in situations where the predictive family $\mathcal{V}$ is not restricted to discretized continuous distributions. However, common inputs for transformers such as images and text are discrete in nature. As such, by the Data Processing Inequality \cite{Cover}, the entropy of their transformed (target) representation cannot be higher, as long as the transformations are deterministic.

\subsubsection{Differential entropy lower-bound assuming side information as a function of a common ancestor}

We assume that the target representation $Y$ and the side information $W$ originate from the same input $X$ so that the functions generating them are invertible. Due to invertibility, both representations contain all information about the input and each other. We demonstrate that even in this unlikely situation for neural networks, the choice of input transformations can produce an increase in the condition differential entropy $h(Y|W)$.

Before showing the main result of this section, we introduce the following lemma and its proof:

\begin{lemma}
\label{lemma:jacobian-upper-bound}
Given $Y = f(X), W = g(X)$, where $X$ is a continuous random variable of dimensionality $K$, and $f$, $g$ are bijective and Lipschitz continuous with constants $L_f$, $L_g$ respectively, we have:
\begin{align}
    \mathbb{E}_{\mathbf{x} \sim X}
    \left[
    \log
    \frac{
    \abs{J_{g(\mathbf{x})}}
    }{
    \abs{J_{f(\mathbf{x})}}
    }
    \right]
    \leq
    (L_g + L_{f^{-1}} - 2)K
\end{align}
\begin{proof}
\begin{align}
    \mathbb{E}_{\mathbf{x} \sim X}
    \left[
    \log
    \frac{
    \abs{J_{g(\mathbf{x})}}
    }{
    \abs{J_{f(\mathbf{x})}}
    }
    \right]
    &=
    \mathbb{E}_{\mathbf{x} \sim X}
    \left[
    \log
    \frac{
    \abs{\prod_{i=1}^K \lambda_i \left( J_{g(\mathbf{x})}\right)}
    }{
    \abs{\prod_{i=1}^K \lambda_i \left( J_{f(\mathbf{x})}\right)}
    }
    \right]
    \\
    &=
    \mathbb{E}_{\mathbf{x} \sim X}
    \left[
    \sum_{i=1}^K \log \abs{ \lambda_i \left( J_{g(\mathbf{x})}\right)}
    -
    \sum_{i=1}^K \log \abs{ \lambda_i \left( J_{f(\mathbf{x})}\right)}
    \right]
    \\
    &\leq
    \label{eq:log_bounds}
    \mathbb{E}_{\mathbf{x} \sim X}
    \left[
    \sum_{i=1}^K
    \left\{
    \abs{ \lambda_i \left( J_{g(\mathbf{x})}\right)}
    -
    1
    \right\}
    -
    \sum_{i=1}^K
    \left\{
    1
    -
    \frac{
    1
    }{
    \abs{ \lambda_i \left( J_{f(\mathbf{x})}\right)}
    }
    \right\}
    \right]
    \\
    &=
    \mathbb{E}_{\mathbf{x} \sim X}
    \left[
    \sum_{i=1}^K
    \abs{ \lambda_i \left( J_{g(\mathbf{x})}\right)}
    +
    \sum_{i=1}^K
    \frac{
    1
    }{
    \abs{ \lambda_i \left( J_{f(\mathbf{x})}\right)}
    }
    \right]
    -
    2K
    \\
    &=
    \label{eq:eigenvalues}
    \mathbb{E}_{\mathbf{x} \sim X}
    \left[
    \norm{ \bm{\lambda} \left( J_{g(\mathbf{x})}\right) }_1
    +
    \norm{ \bm{\lambda} \left( J^{-1}_{f(\mathbf{x})}\right) }_1
    \right]
    -
    2K
    \\
    \label{eq:norm_1_bound}
    &\leq
    \mathbb{E}_{\mathbf{x} \sim X}
    \left[
    \norm{ \bm{\lambda} \left( J_{g(\mathbf{x})}\right) }_2
    +
    \norm{ \bm{\lambda} \left( J^{-1}_{f(\mathbf{x})}\right) }_2
    \right]
    \sqrt{K}
    -
    2K
    \\
    \label{eq:norm_2_bound}
    &=
    \mathbb{E}_{\mathbf{x} \sim X}
    \left[
    \norm{ J_{g(\mathbf{x})} }_{\mathrm{F}}
    +
    \norm{ J^{-1}_{f(\mathbf{x})} }_{\mathrm{F}}
    \right]
    \sqrt{K}
    -
    2K
    \\
    \label{eq:norm_f_bound}
    &\leq
    \mathbb{E}_{\mathbf{x} \sim X}
    \left[
    \norm{ J_{g(\mathbf{x})} }_2
    +
    \norm{ J^{-1}_{f(\mathbf{x})} }_2
    \right]
    K
    -
    2K
    \\
    \label{eq:jacobians}
    &=
    \mathbb{E}_{\mathbf{x} \sim X}
    \left[
    \norm{ J_{g(\mathbf{x})} }_2
    +
    \norm{ J_{f^{-1}(f(\mathbf{x}))} }_2
    \right]
    K
    -
    2K
    \\
    &\leq
    \label{eq:lip_norm_bound}
    (L_g + L_{f^{-1}} - 2)K
    .
\end{align}
Eq.~\ref{eq:log_bounds} uses the logarithmic bounds
$
\log a \leq a - 1;
\log a \geq 1 - \nicefrac{1}{a}
$.
Eq.~\ref{eq:eigenvalues} uses the fact that $
\forall \lambda_i(A) \in \bm{\lambda}(A)
\; \exists \,
\nicefrac{1}{\lambda_i(A)} \in \bm{\lambda}(A^{-1})
$.
Eq.~\ref{eq:norm_1_bound} uses the bound for the 1-norm
$
\norm{\mathbf{a}}_1
\leq
\sqrt{\abs{\mathbf{a}}} \, \norm{\mathbf{a}}_2
$.
Eq.~\ref{eq:norm_2_bound} uses the bound for the 2-norm
$
\norm{\bm{\lambda}(A)}_2
=
\norm{A}_{\mathrm{F}}
$.
Eq.~\ref{eq:norm_f_bound} uses the Frobenius norm bound
$
\norm{A}_{\mathrm{F}}
\leq
\sqrt{\mathrm{rank}(A)}
\,
\norm{A}_2
$.
Eq.~\ref{eq:jacobians} uses the relationship between Jacobian inverses
$\smash{J^{-1}_{f(\mathbf{x})} = J_{f^{-1}(f(\mathbf{x}))}}$.
Finally, Eq.~\ref{eq:lip_norm_bound} uses the Lipschitz continuity assumption of $f$ and $g$ and the Jacobian 2-norm bound
$
\norm{J_{f(\mathbf{x})}}_2 \leq L_f \; \forall \mathbf{x} \in \mathbb{R}^K
$.
\end{proof}
\end{lemma}

\begin{theorem}
\label{theorem:ca}
Let $Y = f(X), W = g(X)$, where $X$, $Y$, and $W$ are continuous random variables with dimensionality $K$, and $f$, $g$ are bijective and Lipschitz continuous with constants $L_f$ and $L_g$, respectively. Then, we have:
\begin{equation*}
    h(Y|W)
    \geq
    (2 - L_g - L_{f^{-1}})K
    ,
\end{equation*}
where $L_{f^{-1}}$ is the Lipschitz constant of the inverse of $f$.

\begin{proof}
Since random variables $Y$ and $W$ share a common ancestor $X$, using the change of variables, their probabilities are related by:
\begin{align}
    \label{eq:covc}
    f_{Y}(f(\mathbf{x}))
    =
    f_{W}(g(\mathbf{x}))
    \frac{
    \abs{J_{g(\mathbf{x})}}
    }{
    \abs{J_{f(\mathbf{x})}}
    }
    ,
\end{align}
where $\abs{J_{v(\mathbf{x)}}}$ is the absolute Jacobian determinant of $v(\mathbf{x})$. Thus:
\begin{align}
    h(Y|W)
    &\triangleq
    \mathbb{E}_{\mathbf{y},\mathbf{w} \sim Y,W}
    \left[
    -
    \log
    f_{Y|W}(\mathbf{y}|\mathbf{w})
    \right]
    \\
    \label{eq:bayes}
    &=
    \mathbb{E}_{\mathbf{y},\mathbf{w} \sim Y,W}
    \left[
    -
    \log
    \frac{
    f_{W|Y}(\mathbf{w}|\mathbf{y})
    f_{Y}(\mathbf{y})
    }{
    f_{W}(\mathbf{w})
    }
    \right]
    \\
    &=
    \mathbb{E}_{\mathbf{y},\mathbf{w} \sim Y,W}
    \left[
    -
    \log
    f_{W|Y}(\mathbf{w}|\mathbf{y})
    \right]
    -
    \mathbb{E}_{\mathbf{y},\mathbf{w} \sim Y,W}
    \left[
    \log
    \frac{
    f_{Y}(\mathbf{y})
    }{
    f_{W}(\mathbf{w})
    }
    \right]
    \\
    \label{eq:degenerate}
    &=
    -
    \mathbb{E}_{\mathbf{y},\mathbf{w} \sim Y,W}
    \left[
    \log
    \frac{
    f_{Y}(\mathbf{y})
    }{
    f_{W}(\mathbf{w})
    }
    \right]
    \\
    \label{eq:covc_use}
    &=
    -
    \mathbb{E}_{\mathbf{x} \sim X}
    \left[
    \mathbb{E}_{\mathbf{y} \sim Y|g(\mathbf{x})}
    \left[
    \log
    \frac{
    f_{Y}(\mathbf{y})
    }{
    f_{W}(g(\mathbf{x}))
    }
    \right]
    \right]
    \\
    &=
    -
    \mathbb{E}_{\mathbf{x} \sim X}
    \left[
    \log
    \frac{
    \abs{J_{g(\mathbf{x})}}
    }{
    \abs{J_{f(\mathbf{x})}}
    }
    \right]
    \\
    \label{eq:lemma_lip}
    &\geq
    (2 - L_g - L_{f^{-1}})K
    .
\end{align}
Eq.~\ref{eq:bayes} uses Bayes' theorem. 
Eq.~\ref{eq:degenerate} is due to $P_{W|Y}$ being degenerate since $f$ and $g$ are bijective. 
Eq.~\ref{eq:covc_use} uses Eq.~\ref{eq:covc}. 
Eq.~\ref{eq:lemma_lip} uses Lemma~\ref{lemma:jacobian-upper-bound}.
\end{proof}
\end{theorem}

We see that the differential entropy decreases as $g$ expands and $f$ contracts. The second term cancels when $f$ and $g$ are identity functions.

\subsubsection{Differential entropy lower-bound assuming side information as a function of the target representation}

This formulation offers an alternative view in cases where the side information is considered fixed, such that the hyper-prior analysis transform $g_h$ is not part of the predictive family.

\begin{theorem}
\label{theorem:fixed}
Let $\smash{Y = f(X), W = g(Y)}$, where $X$, $Y$, and $W$ are continuous random variables with dimensionality $K$, $f$ and $g$ are bijective, and $g$ is Lipschitz continuous with constant $L_g$. Then, we have: 
\begin{align}
    h(Y|W)
    \geq
    (1 - L_g)K
    .
\end{align}

\begin{proof}
Following the proof for Lemma~\ref{lemma:jacobian-upper-bound} closely, we arrive at:
\begin{align}
    \mathbb{E}_{\mathbf{x} \sim X}
    \left[
    \log
    \frac{
    \abs{J_{(g \circ f)(\mathbf{x})}}
    }{
    \abs{J_{f(\mathbf{x})}}
    }
    \right]
    &=
    \mathbb{E}_{\mathbf{x} \sim X}
    \left[
    \log
    \frac{
    \abs{J_{g(f(\mathbf{x}))} J_{f(\mathbf{x})}}
    }{
    \abs{J_{f(\mathbf{x})}}
    }
    \right]
    \\
    \label{eq:log-jac}
    &=
    \mathbb{E}_{\mathbf{x} \sim X}
    \left[
    \log
    \abs{J_{g(f(\mathbf{x}))}}
    \right]
    \\
    &=
    \mathbb{E}_{\mathbf{x} \sim X}
    \left[
    \log
    \abs{\prod_{i=1}^K \lambda_i \left( J_{g(f(\mathbf{x}))}\right)}
    \right]
    \\
    &\leq
    \mathbb{E}_{\mathbf{x} \sim X}
    \left[
    \sum_{i=1}^K
    \abs{ \lambda_i \left( J_{g(f(\mathbf{x}))}\right)}
    \right]
    -
    K
    \\
    &\leq
    \mathbb{E}_{\mathbf{x} \sim X}
    \left[
    \norm{ J_{g(f(\mathbf{x}))} }_2
    \right]
    K
    -
    K
    \\
    \label{eq:lips-from-jac}
    &\leq
    (L_g - 1) K
    .
\end{align}
Plugging this result into Eq.~\ref{eq:covc_use} and changing variables with respect to $X$ arrives at the result.
\end{proof}
\end{theorem}

We see that as $g$ contracts, the differential entropy increases. The term cancels when $g$ becomes the identity function. 

\subsection{Generalization bounds for \texorpdfstring{$\mathcal{V}$}{V}-entropy and the gap}

\subsubsection{Revisiting the generalization bound for \texorpdfstring{$\mathcal{V}$}{V}-entropy}

\getkeytheorem{theorem:v-gen-bound}
\begin{proof}
Starting from Lemma 3 in \cite{XuZSSE20}, we have:
\begin{align}
    R_{\mathcal{V},\mathcal{D}}(Y|W)
    &\leq
    2
    \mathfrak{R}(\mathcal{V}_r \circ \mathcal{D})
    +
    B
    \sqrt{
    \nicefrac{2}{N}
    \log \nicefrac{1}{\delta}
    }
    \\
    \label{eq:talengrad-first}
    &\leq
    2
    \mathfrak{L}(\mathcal{V}_r)
    \,
    \mathfrak{R}(\mathcal{D})
    +
    B
    \sqrt{
    \nicefrac{2}{N}
    \log \nicefrac{1}{\delta}
    }
    .
\end{align}
Eq.~\ref{eq:talengrad-first} uses the Kakade \& Tewari Lemma \cite{KakadeTewari} based on Talagrand’s contraction principle \cite{ledoux2013probability, BartlettM02}. It states that if all vectors in a set $A$ are operated by a Lipschitz function, then $\mathfrak{R}(A)$ is at most multiplied by the Lipschitz constant of the function.
\end{proof}

We provide a generalization error bound for $\mathcal{R}_{\mathcal{V}, \mathcal{D}}(Y)$ (i.e. with no side information) when the predictive family $\mathcal{V}$ is composed of discretized multivariate normal distributions with diagonal covariances. First, we upper-bound the Lipschitz constant of the logarithm of such predictive family:

\begin{lemma}
\label{lemma:lip-gauss}
Let $Y$ be a random variable with sample space $\mathcal{Y}$, and $\mathcal{V} \subseteq \Omega$ be a predictive family \cite{XuZSSE20} of discretized multivariate normal distributions with diagonal covariances lower-bounded by $\sigma^2_{\mathrm{min}}$, where $\forall g \in \mathcal{V}, \mathbf{y} \in \mathcal{Y}, \log g[\oslash](\mathbf{y}) \in [-B, B]$. We have that:
\begin{align}
    \mathfrak{L}(\mathcal{V}_r)
    \leq
    e^{B}
    \sqrt{2 \pi}
    \sigma_{\mathrm{min}},
\end{align}
where $\mathfrak{L}(\mathcal{V}_r)$ is the maximum Lipschitz constant in $\mathcal{V}_r = \{v| v(\mathbf{y}) = \log g[\oslash](\mathbf{y}), g \in \mathcal{V} \}$.

\begin{proof}
With $\Delta$ as a fixed-step size for discretization, and $\smash{\Sigma \in [\sigma^2_{\mathrm{min}}, \infty)^K}$, we have:
\begin{align}
    \mathfrak{L}(\mathcal{V}_r)
    \label{eq:lip-def}
    &\triangleq
    \sup_{
    g \in \mathcal{V}_r
    ,
    \mathbf{y} \in \mathcal{Y}
    }
    \norm{
    \nabla
    \log
    g[\oslash](\mathbf{y})
    }_\infty
    \\
    &\leq
    \sup_{
    \bm\mu \in \mathbb{R}^K
    ,
    \bm\sigma^2 \in \Sigma
    ,
    \mathbf{y} \in \mathcal{Y}
    }
    \norm{
    \nabla
    \log
    \prod_{i=1}^K
    \left[
    \Phi(y_i + \Delta; \mu_i, \sigma^2_i)
    -
    \Phi(y_i - \Delta; \mu_i, \sigma^2_i)
    \right]
    }_\infty
    \\
    &=
    \sup_{
    \bm\mu \in \mathbb{R}^K
    ,
    \bm\sigma^2 \in \Sigma
    ,
    \mathbf{y} \in \mathcal{Y}
    }
    \norm{
    \sum_{i=1}^K
    \nabla
    \log
    \left[
    \Phi(y_i + \Delta; \mu_i, \sigma^2_i)
    -
    \Phi(y_i - \Delta; \mu_i, \sigma^2_i)
    \right]
    }_\infty
    \\
    &=
    \sup_{
    \bm\mu \in \mathbb{R}^K
    ,
    \bm\sigma^2 \in \Sigma
    ,
    \mathbf{y} \in \mathcal{Y}
    }
    \max_{i=1}^K
    \abs{
    \frac{\partial}{\partial y_i}
    \log
    \left[
    \Phi(y_i + \Delta; \mu_i, \sigma^2_i)
    -
    \Phi(y_i - \Delta; \mu_i, \sigma^2_i)
    \right]
    }
    \\
    &=
    \sup_{
    \bm\mu \in \mathbb{R}^K
    ,
    \bm\sigma^2 \in \Sigma
    ,
    \mathbf{y} \in \mathcal{Y}
    }
    \max_{i=1}^K
    \frac{
    \abs{
    \mathcal{N}(y_i + \Delta; \mu_i, \sigma^2_i)
    -
    \mathcal{N}(y_i - \Delta; \mu_i, \sigma^2_i)
    }
    }{
    \Phi(y_i + \Delta; \mu_i, \sigma^2_i)
    -
    \Phi(y_i - \Delta; \mu_i, \sigma^2_i)
    }
    \\
    \label{eq:use-b-bound}
    &\leq
    e^B
    \sup_{
    \bm\mu \in \mathbb{R}^K
    ,
    \bm\sigma^2 \in \Sigma
    ,
    \mathbf{y} \in \mathcal{Y}
    }
    \max_{i=1}^K
    \abs{
    \mathcal{N}(y_i + \Delta; \mu_i, \sigma^2_i)
    -
    \mathcal{N}(y_i - \Delta; \mu_i, \sigma^2_i)
    }
    \\
    \label{eq:use-gauss-prop}
    &\leq
    e^B
    \sup_{
    \bm\mu \in \mathbb{R}^K
    ,
    \bm\sigma^2 \in \Sigma
    ,
    \mathbf{y} \in \mathcal{Y}
    }
    \max_{i=1}^K
    \mathcal{N}(y_i + \Delta; y_i + \Delta, \sigma^2_i)
    \\
    &=
    e^{B}
    \sup_{
    \sigma^2 \in [\sigma^2_{\mathrm{min}}, \infty)
    }
    \mathcal{N}(0; 0, \sigma^2)
    \\
    &=
    e^{B}
    \sup_{
    \sigma^2 \in [\sigma^2_{\mathrm{min}}, \infty)
    }
    (2 \pi \sigma^2)^{-\nicefrac{1}{2}}
    \\
    &\leq
    e^{B}
    (2 \pi \sigma^2_{\mathrm{min}})^{-\nicefrac{1}{2}}
    .
\end{align}
Eq.~\ref{eq:lip-def} uses a definition of Lipschitz constant \cite{benyaminigeometric}. Eq.~\ref{eq:use-b-bound} uses the bound $\log g[\oslash](\mathbf{y}) \geq -B$. Eq.~\ref{eq:use-gauss-prop} uses $\mathcal{N}(\cdot) \geq 0$ and the fact that the normal probability density function is the highest at the mean.
\end{proof}
\end{lemma}

Finally, we plug this result in the generalization error bound:

\begin{corollary}
\label{corollary:gen-bound-gaussian}
Let $Y$ be a random variable with sample space $\mathcal{Y}$, $\mathcal{D} = \{ \mathbf{y}_i \}_{i=1}^N \sim Y$ be a set of its samples, and $\mathcal{V} \subseteq \Omega$ be a predictive family \cite{XuZSSE20} of discretized multivariate normal distributions with diagonal covariances lower-bounded by $\sigma^2_{\mathrm{min}}$, where $\forall g \in \mathcal{V}, \mathbf{y} \in \mathcal{Y}, \log g[\oslash](\mathbf{y}) \in [-B, B]$. We have that:
\begin{align}
    R_{\mathcal{V},\mathcal{D}}(Y)
    &\leq
    2
    e^{B}
    (2 \pi \sigma^2_{\mathrm{min}})^{-\nicefrac{1}{2}}
    \,
    \mathfrak{R}(\mathcal{D})
    +
    B
    \sqrt{
    \nicefrac{2}{N}
    \log \nicefrac{1}{\delta}
    }
    .
\end{align}

\begin{proof}
Use Lemma~\ref{lemma:lip-gauss} in Theorem~\ref*{theorem:v-gen-bound}.
\end{proof}
\end{corollary}

\subsubsection{Generalization bound for the \texorpdfstring{$\mathcal{V}$}{V}-entropy gap}

We define the generalization error for the $\mathcal{V}$-entropy gap as:

\begin{definition}
\label{def:gen}
Let $\mathcal{D} = \{(\mathbf{y}_i, \mathbf{w}_i) \}_{i=1}^N \sim Y, W$ be a set of samples. We define the generalization error of the $\mathcal{V}$-entropy gap as:
\begin{align*}
    S_{\mathcal{V},\mathcal{D}}(Y|W)
    \triangleq
    \Bigg\vert
    G_{\mathcal{V}}(Y|W)
    -
    \inf_{g \in \mathcal{V}}
    \frac{1}{N}
    \sum_{(\mathbf{y},\mathbf{w}) \in \mathcal{D}}
    \log
    \frac{
    P_{Y|W}(\mathbf{y}|\mathbf{w})
    }{
    g[\mathbf{w}](\mathbf{y})
    }
    \Bigg\vert
    .
\end{align*}
\end{definition}

This term can be upper-bounded in terms of the Rademacher complexity \cite{LearningTheory} of the target representation, the Lipschitz constants of the predictive family $\mathcal{V}$ and of $\log P_{Y|W}$, the log of the true conditional probability function for the target representation:

\begin{theorem}
\label{theorem:gen-bound-gap}
Let $\mathcal{V} \subseteq \Omega$ be a predictive family \cite{XuZSSE20}, $Y$ and $W$ be random variables with sample spaces $\mathcal{Y}$ and $\mathcal{W}$, respectively, and $\mathcal{D} = \{(\mathbf{y}_i, \mathbf{w}_i) \}_{i=1}^N \sim Y, W$ be a set of their samples. Assume that $\forall g \in \mathcal{V}, \mathbf{y} \in \mathcal{Y}, \mathbf{w} \in \mathcal{W}, \log g[\mathbf{w}](\mathbf{y}) \in [-B, B], \log P_{Y|W}(\mathbf{y}|\mathbf{w}) \in [-B, B]$, and that all these probability functions are Lipschitz continuous. Then, $\forall \delta \in (0, 1)$, with probability at least $1 - \delta$, we have:
\begin{equation*}
    S_{\mathcal{V},\mathcal{D}}(Y|W)
    \leq
    2
    \left\{
    \left[
    \mathfrak{L}(\mathcal{V}_r)
    +
    \mathfrak{L}(\log P_{Y|W})
    \right]
    \mathfrak{R}(\mathcal{D})
    +
    B
    \sqrt{
    \nicefrac{2}{N}
    \log \nicefrac{1}{\delta}
    }
    \right\}
    ,
\end{equation*}
where $\mathfrak{R}(\mathcal{D})$ is the Rademacher complexity of the concatenated samples in the dataset $\mathcal{D}$, $\mathfrak{L}(\mathcal{V}_r)$ is the maximum Lipschitz constant in $\mathcal{V}_r = \{v| v(\mathbf{w}, \mathbf{y}) = \log g[\mathbf{w}](\mathbf{y}), g \in \mathcal{V} \}$, and $\mathfrak{L}(\log P_{Y|W})$ is the largest Lipschitz constant in $\{\log P_{Y|\mathbf{w}}, \mathbf{w} \in \mathcal{W}\}$.

\begin{proof}
With $\hat{g} = \argmin_{g \in \mathcal{V}} \sum_{(\mathbf{y}, \mathbf{w}) \in \mathcal{D}} - \log g[\mathbf{w}](\mathbf{y})$, we derive: 
\begin{align}
    S_{\mathcal{V},\mathcal{D}}(Y|W)
    &\triangleq
    \abs{
    G_{\mathcal{V}}(Y|W)
    -
    \inf_{g \in \mathcal{V}}
    \frac{1}{N}
    \sum_{(\mathbf{y}, \mathbf{w}) \in \mathcal{D}}
    \log
    \frac{
    P_{Y|W}(\mathbf{y}|\mathbf{w})
    }{
    g[\mathbf{w}](\mathbf{y})
    }
    }
    \\
    &=
    \abs{
    H_{\mathcal{V}}(Y|W)
    -
    H_{\Omega}(Y|W)
    -
    \frac{1}{N}
    \sum_{(\mathbf{y}, \mathbf{w}) \in \mathcal{D}}
    \log
    \frac{
    P_{Y|W}(\mathbf{y}| \mathbf{w})
    }{
    \hat{g}[\mathbf{w}](\mathbf{y})
    }
    }
    \\
    \label{eq:triangle}
    &\leq
    \abs{
    H_{\mathcal{V}}(Y|W)
    -
    \frac{1}{N}
    \sum_{(\mathbf{y}, \mathbf{w}) \in \mathcal{D}}
    \mkern-14mu
    -
    \log
    \hat{g}[\mathbf{w}](\mathbf{y})
    }
    +
    \abs{
    \frac{1}{N}
    \sum_{(\mathbf{y}, \mathbf{w}) \in \mathcal{D}}
    \mkern-14mu
    -
    \log
    P_{Y|W}(\mathbf{y} | \mathbf{w})
    -
    H_{\Omega}(Y|W)
    }
    \\
    \label{eq:pac}
    &\leq
    2
    \,
    \mathfrak{R}(\mathcal{V}_r \circ \mathcal{D})
    +
    2
    \,
    \mathfrak{R}(\log P_{Y|W} \circ \mathcal{D})
    +
    2B
    \sqrt{
    \nicefrac{2}{N}
    \log \nicefrac{1}{\delta}
    }
    \\
    \label{eq:talengrad}
    &\leq
    2
    \,
    \mathfrak{L}(\mathcal{V}_r)
    \,
    \mathfrak{R}(\mathcal{D})
    +
    2
    \,
    \mathfrak{L}(\log P_{Y|W})
    \,
    \mathfrak{R}(\mathcal{D})
    +
    2B
    \sqrt{
    \nicefrac{2}{N}
    \log \nicefrac{1}{\delta}
    }
    \\
    &=
    2
    \left\{
    \left[
    \mathfrak{L}(\mathcal{V}_r)
    +
    \mathfrak{L}(\log P_{Y|W})
    \right]
    \mathfrak{R}(\mathcal{D})
    +
    B
    \sqrt{
    \nicefrac{2}{N}
    \log \nicefrac{1}{\delta}
    }
    \right\}
    .
\end{align}
Eq.~\ref{eq:triangle} uses the triangle inequality. Eq.~\ref{eq:pac} uses Lemma 3 in \cite{XuZSSE20} on each of the two terms, where the predictive family for the second term has been reduced to $\smash{\{ P_{Y|W} \}}$. Eq.~\ref{eq:talengrad} uses the Kakade \& Tewari Lemma \cite{KakadeTewari} based on Talagrand’s contraction principle \cite{ledoux2013probability, BartlettM02}. It states that if all vectors in a set $A$ are operated by a Lipschitz function, then $\mathfrak{R}(A)$ is at most multiplied by the Lipschitz constant of the function.
\end{proof}
\end{theorem}

Unlike its $\mathcal{V}$-entropy counterpart, this bound has the additional term that captures the difficulty of estimating the entropy of a target representation $Y$. This complexity is bounded by the Lipschitz constant of the log of its conditional probability functions, $\mathfrak{L}(\log P_{Y|W})$, assuming they are Lipschitz continuous. For $\smash{R_{\mathcal{V},\mathcal{D}}(Y|Y)}$, where a potentially rate-constrained predictive family $\mathcal{V}$ is used, as is the case in this work, we have $\mathfrak{L}(\log P_{Y|W}) = 0$, which nullifies the impact of this complexity.

\section{Additional experimental results and details}

\subsection{Entropy model definitions}
Let $\mathcal{F}(\mathbb{R})$ be the set of all \textit{cumulative density functions} (CDFs), and  $\Theta = \{ \bm{\theta}_1,...,\bm{\theta}_C \}$ be a set of parameter vectors for each embedding dimension of $W$. A zero-context learned entropy model for the hyper-prior $g_w : \Theta \to \mathcal{F}(\mathbb{R})$ takes a parameter vector $\bm{\theta}_j; j = \{1,...,C\}$ to generate a CDF for any element in the $j$-th embedding dimension of $W$. The rate of a hyper-prior $\mathbf{w} \sim W$ is the fully-factorized negative log-likelihood of a unit interval centered around $\mathbf{w}$:
\begin{equation}
    \label{eq:hp-rate}
    r_w(\mathbf{w})
    =
    -
    \sum_{i=1}^T
    \sum_{j=1}^C
    \log
    \Big[
    g_w
    [\bm{\theta}_j]
    \left(
    w_{i,j} + \nicefrac{1}{2}
    \right)
    \\
    -
    g_w
    [\bm{\theta}_j]
    \left(
    w_{i,j} - \nicefrac{1}{2}
    \right)
    \Big]
    .
\end{equation}

Assuming a conditionally independent distribution for $Y$ given $W$, the rate for $\smash{\mathbf{y} \sim Y}$ is given by the negative log-likelihood of a unit interval centered around $\mathbf{y}$:
\begin{align}
    \label{eq:rate}
    r_y(\mathbf{y}; \mathbf{w})
    =
    - 
    \sum_{i=1}^{T \times E}
    \log
    \Big[
    \Phi
    \left(
    y_i + \nicefrac{1}{2}
    ;\;
    g_\mu(\mathbf{w})_i
    ,\,
    g_\sigma(\mathbf{w})_i
    \right)
    -
    \Phi
    \left(
    y_i - \nicefrac{1}{2}
    ;\;
    g_\mu(\mathbf{w})_i
    ,\,
    g_\sigma(\mathbf{w})_i
    \right)
    \Big]
    ,
\end{align}
where $\smash{g_\mu : \mathcal{W} \to \mathbb{R}^{T \times E}}$ and $\smash{g_\sigma : \mathcal{W} \to \mathbb{R}_{+}^{T \times E}}$ correspond to the means and variances, respectively, produced by the entropy model $g_y$, $\Phi$ is the normal CDF, and $i$ indexes the elements in the tensors.

\subsection{Result breakdown}

\begin{table}[!t]
\centering
\caption{GPT-2 rate-distortion performance}
\label{table:gpt2}
\begin{tabular}{lcrrrrr}
    \toprule
    Model & Split Point & $\lambda$ & Hyper-Prior BPT & Total BPT & Perplexity & LAMBADA \\
    \midrule
    Deep factorized & 3 & 0.0050 & 32.56 & 155.94 & 20.64 & 0.2713 \\
    Deep factorized & 3 & 0.0100 & 31.91 & 144.37 & 20.76 & 0.2657 \\
    Deep factorized & 3 & 0.0500 & 32.05 & 133.18 & 21.35 & 0.2540 \\
    \midrule
    Direct access & 3 & 0.0010 & 32.91 & 263.94 & 20.85 & 0.2791 \\
    Direct access & 3 & 0.0025 & 30.22 & 198.84 & 21.20 & 0.2583 \\
    Direct access & 3 & 0.0075 & 26.84 & 150.94 & 22.07 & 0.2655 \\
    Direct access & 3 & 0.0100 & 29.88 & 144.36 & 22.31 & 0.2604 \\
    \midrule
    Deep factorized & 6 & 0.0010 & 45.55 & 355.93 & 20.89 & 0.2824 \\
    Deep factorized & 6 & 0.0025 & 40.48 & 273.06 & 21.25 & 0.2723 \\
    Deep factorized & 6 & 0.0075 & 37.69 & 227.85 & 22.00 & 0.2606 \\
    Deep factorized & 6 & 0.0100 & 36.70 & 217.08 & 22.30 & 0.2525 \\
    \midrule
    Direct access & 6 & 0.0010 & 32.63 & 268.52 & 22.23 & 0.2554 \\
    Direct access & 6 & 0.0025 & 31.11 & 216.90 & 22.59 & 0.2663 \\
    Direct access & 6 & 0.0075 & 30.47 & 165.94 & 25.22 & 0.2296 \\
    Direct access & 6 & 0.0100 & 30.24 & 160.16 & 26.24 & 0.2084 \\
    \midrule
    Fourier basis & 6 & 0.001 & 249.18 & 541.67 & 21.15 & 0.2946 \\
    Fourier basis & 6 & 0.025 & 248.14 & 464.49 & 21.70 & 0.2800 \\
    Fourier basis & 6 & 0.075 & 309.20 & 434.07 & 24.03 & 0.2212 \\
    Fourier basis & 6 & 0.010 & 302.62 & 422.32 & 24.19 & 0.2069 \\
    \midrule
    Deep factorized & 9 & 0.0010 & 57.09 & 432.86 & 20.97 & 0.2864 \\
    Deep factorized & 9 & 0.0025 & 43.24 & 315.58 & 22.07 & 0.2659 \\
    Deep factorized & 9 & 0.0050 & 41.11 & 291.92 & 22.30 & 0.2482 \\
    Deep factorized & 9 & 0.0075 & 40.69 & 279.71 & 22.82 & 0.2529 \\
    \midrule
    Direct access & 9 & 0.0008 & 39.27 & 419.81 & 22.26 & 0.2614 \\
    Direct access & 9 & 0.0025 & 36.20 & 322.79 & 23.55 & 0.2525 \\
    Direct access & 9 & 0.0075 & 29.10 & 251.55 & 24.84 & 0.2117 \\
    Direct access & 9 & 0.0100 & 29.05 & 241.31 & 25.52 & 0.2007 \\
    \midrule
    Uncompressed & Any & -- & -- & 12,288 & 20.71 & 0.2961 \\
    \bottomrule
\end{tabular}
\end{table}

\begin{table}[!t]
\centering
\caption{Split point rate-distortion performance for language models}
\label{table:lm}
\begin{tabular}{lcrrrrrr}
    \toprule
    Model & Split Point & $\lambda$ & BPT & Perplexity & $\mathfrak{R}(\mathcal{D})$ & $\log \mathfrak{L}(v)$ & $\nicefrac{1}{2D} \log \abs{\Sigma}$ \\
    \midrule
    GPT-2 & 3 & 0.010 & 148.7 & 19.10 & 1.18 $\pm$ 0.007 & 6.62 $\pm$ 0.097 & 4.57 $\pm$ 0.002 \\
    GPT-2 & 6 & 0.001 & 355.9 & 20.89 & 5.27 $\pm$ 0.052 & 4.21 $\pm$ 0.036 & 4.90 $\pm$ 0.002 \\
    GPT-2 & 9 & 0.001 & 492.1 & 21.61 & 13.53 $\pm$ 0.100 & 5.04 $\pm$ 0.107 & 5.35 $\pm$ 0.002 \\
    \midrule
    Pythia & 3 & 0.001 & 371.5 & 21.36 & 9.95 $\pm$ 0.045 & 5.03 $\pm$ 0.046 & 5.17 $\pm$ 0.001 \\
    Pythia & 6 & 0.001 & 451.9 & 21.99 & 12.12 $\pm$ 0.048 & 5.78 $\pm$ 0.044 & 5.26 $\pm$ 0.002 \\
    Pythia & 8 & 0.001 & 513.6 & 22.39 & 14.39 $\pm$ 0.108 & 5.42 $\pm$ 0.033 & 5.26 $\pm$ 0.002 \\
    \bottomrule
\end{tabular}
\end{table}

\begin{table}[!t]
\centering
\caption{Split point rate-distortion performance for image classification}
\label{table:ic}
\begin{tabular}{lcrrrrrr}
    \toprule
    Model & Split Point & $\lambda$ & BPP & Accuracy & $\mathfrak{R}(\mathcal{D})$ & $\log \mathfrak{L}(v)$ & $\nicefrac{1}{2D} \log \abs{\Sigma}$ \\
    \midrule
    ViT & 3 & 0.01 & 2.35 & 0.79 & 0.12 $\pm$ 0.000 & 4.20 $\pm$ 0.011 & 3.99 $\pm$ 0.002 \\
    ViT & 6 & 0.01 & 3.27 & 0.79 & 0.31 $\pm$ 0.001 & 4.16 $\pm$ 0.003 & 4.35 $\pm$ 0.003 \\
    ViT & 9 & 0.01 & 5.15 & 0.79 & 0.56 $\pm$ 0.002 & 2.47 $\pm$ 0.006 & 5.53 $\pm$ 0.002 \\
    \midrule
    ResNet & 3 & 0.01 & 4.75 & 0.68 & 1.87 $\pm$ 0.014 & 4.13 $\pm$ 0.008 & 5.93 $\pm$ 0.002 \\
    ResNet & 7 & 0.01 & 2.65 & 0.67 & 1.12 $\pm$ 0.004 & 4.14 $\pm$ 0.009 & 5.76 $\pm$ 0.001 \\
    ResNet & 13 & 0.01 & 1.20 & 0.67 & 1.03 $\pm$ 0.004 & 4.02 $\pm$ 0.018 & 5.38 $\pm$ 0.001 \\
    \bottomrule
\end{tabular}
\end{table}

\begin{table}[!t]
\centering
\caption{Hyper-parameter settings}
\label{table:parameters}
\begin{tabular}{lrrrr}
    \toprule
    Parameter & GPT-2 & Pythia & ViT & ResNet \\
    \midrule
    Precision & bfloat16 & bfloat16 & bfloat16 & float32 \\
    Target representation dimensionality ($E$) & 768 & 768 & 768 & 768 \\
    Side information dimensionality ($C$) & 24 & 24 & 24 & 24 \\
    Target representation context size ($T$) & 1,024 & 1,024 & 49 & 49 \\
    Side information density parameters ($\smash{\abs{\bm{\theta}_j}}$) & 118 & 118 & 118 & 118 \\
    \midrule
    Distortion function & cross-entropy & cross-entropy & cross-entropy & cross-entropy \\
    Tokenizer & GPT-2 & GPT-2 & $-$ & $-$ \\
    Label smoothing & $-$ & $-$ & 0.11 & 0 \\
    MixUp coefficient & $-$ & $-$ & 0.2 & 0.2 \\
    CutMix coefficient & $-$ & $-$ & 1 & 1 \\
    Random augmentations & $-$ & $-$ & 2 & 2 \\
    Augmentation magnitude & $-$ & $-$ & 9 & 9 \\
    \midrule
    Batch size & 12 & 12 & 16 & 16 \\
    Accumulated gradient batches & 40 & 40 & 40 & 40 \\
    \midrule
    Optimizer & AdamW & AdamW & AdamW & AdamW \\
    Optimizer parameters & (0.9, 0.95) & (0.9, 0.95) & (0.9, 0.95) & (0.9, 0.95) \\
    Maximum learning rate & 0.0006 & 0.0006 & 0.0006 & 0.0006 \\
    Minimum learning rate & 0.00006 & 0.00006 & 0.00006 & 0.00006 \\
    Weight decay & 0.1 & 0.1 & 0.3 & 0.001 \\
    Patience & 5 & 5 & 5 & 5 \\
    Gradient norm clipping & 1 & 1 & 1 & 1 \\
    \midrule
    Warm up function & linear & linear & linear & linear \\
    Warm up steps & 2,000 & 2,000 & 2,000 & 2,000 \\
    Schedule function & cosine & cosine & cosine & cosine \\
    Maximum steps & 600,000 & 600,000 & 600,000 & 600,000 \\
    \bottomrule
\end{tabular}
\end{table}

Tables \ref{table:gpt2}, \ref{table:lm}, and \ref{table:ic} provide the task performances achieved for the models and the corresponding tasks evaluated in \cref*{sec:rd-benchmarks,sec:rd-split-exp}. Confidence intervals for the estimates of the Rademacher complexity of the analysis transform, the covariance determinant of the target representation, and the Lipschitz constant of the entropy model are provided at a confidence level of 99\%. Different randomly-selected subsets of the validation dataset are used for each sample. Any source of randomness used in the computation of the estimates is also different between samples. The underlying models are the same across samples.

\subsection{Experimental settings and resource details}
\label{sec:parameters}

The text is processed using the TikToken GPT-2 tokenizer \cite{GPT2} and each sample has $T = 1024$ tokens, where different documents in the sequence are separated by a special token. Following \cite{Karpathy2022}, we use a linear warmup followed by a cosine learning rate schedule for AdamW \cite{AdamW}, with coefficients $\beta_1=0.9, \beta_2=0.95$, and a weight decay of 0.1 placed on the two-dimensional parameter tensors of the language model. Using gradient accumulation, each optimization step uses 480 samples.

All models were trained on a single NVIDIA A40 GPU. During training, models required at most 20 GB of VRAM. All models were trained with bfloat16 precision except for the ones using the ResNet architecture. The estimates of Rademacher complexities, Lipschitz constants, and covariance determinants were computed using a single NVIDIA A40 GPU, on less than 5 GB of VRAM, taking at most 1 hour per measure and model. An estimate of 320 GPU hours were spent on preliminary experiments.
  
We use the OpenWebText split provided in \cite{Gokaslan2019OpenWeb}. The dataset is provided under the CC0 license. All results pertaining this dataset are reported on the validation set. The OpenWebText is treated as a contiguous text file and a sample is a random window of text with a context size of 1,024. An epoch is considered to be 1,000 gradient descent steps of 480,000 random samples. The validation set consists of 48,000 random samples.  

The ImageNet-1k \cite{RussakovskyDSKS15} is under a custom non-commercial license. Tasks trained on the dataset use the original splits. Each epoch trains on a subset of 100,000 random samples from the training set. All pertaining results are reported on the validation set. Random augmentations such as shearing, translation, rotation, and color jittering are applied to the samples, and MixUp and CutMix transformations are applied to a batch of size 16.  

To produce the rate-distortion curves for the GPT-2 language models, a model is first trained from scratch with $\lambda = 0.0001$ for 50 epochs, or until no improvement has been achieved for more than 5 epochs. This takes around 120 hours. Finally, training is restarted for each $\lambda \in \{0.001, 0.0025, 0.0075, 0.01\}$ from the weights initially obtained. This training is done for around 25 epochs, or until no improvement has been attained for more than 5 epochs. This process usually takes 48 hours. For higher values of $\lambda$ , the loss can diverge during training. In such cases, the maximum learning rate is set to 0.0001 and training is restarted from the checkpoint achieving the lowest loss.  
  
For the Pythia language models, we use $\lambda = 0.001$ . ViT and ResNet use $\lambda=0.01$ on all experiments. The rate used as a loss term for the language models is computed in terms of bits-per-token (BPT), whereas the image classifiers use bits-per-pixel (BPP). The training of the Pythia language models starts from the weights provided in \cite{biderman2023pythia}, under the Apache License, version 2.0. The ViT and ResNet methods use the weights provided in \cite{torchvision2016}, under the BSD 3-Clause license.

Additional hyper-parameters are reported on Table~\ref{table:parameters}.

\subsection{Time benchmarks}

We compare the standalone hyper-prior codec with Deflate \cite{Deutsch1996DEFLATECD} and Zstandard (Zstd) \cite{Collet2018ZstandardCA}, which are lossless data compression algorithms commonly used for tensors. We choose the model with a deep factorized hyper-prior from \cref*{sec:rd-benchmarks} that has the lowest perplexity (highest bitrate) under split point $S=3$. The rate and time produced by the codecs are evaluated on both CPUs and GPUs under heavy parallelization -- as is often the case for inference services -- so that the response throughput is maximized. In the CPU benchmarks, the inference on the hyper-prior used a GPU and the rest of the coding algorithm ran on a CPU.

Deflate uses the LZ77 dictionary-matching compression algorithm and Huffman coding. It is used in the ZIP and PNG file formats. Zstandard combines LZ77 with a large search window and a fast entropy coder. It uses Huffman coding alongside finite-state entropy (FSE), a variant of a tabled asymmetric numeral system (tANS) \cite{Duda2013AsymmetricNS}.

We compare results on the same 1,000 random samples from the OpenWebText validation set. For the GPU benchmarks, we use the NVIDIA nvCOMP \cite{nvCOMP} GPU implementations of GDeflate and Zstandard. For the CPU benchmarks, we use the Python built-in version of Deflate, which is implemented using bindings to the \textit{zlib} C library. For Zstandard, we use the \textit{zstd} Python package, which also uses C bindings to the official implementation. The standalone method uses an implementation of a range asymmetric numeral system (rANS) coder called \textit{torch\_ans} \cite{torchans}.

We run the coding of each token in parallel  from a batch of 10 samples with a context size of $T = 1024$. All benchmarks ran on a NVIDIA H100 SXM5 GPU with 16,896 cores, and 8 cores of an AMD EPYC 9454 CPU.

\begin{table}[!t]
\centering
\caption{GPU codec performance}
\label{table:speed-gpu}
\begin{tabular}{lrrrr}
    \toprule
    Codec & Rate (BPT) & Rate (\%) & Time ($\mu$SPT) & W2W ($\mu$SPT) \\
    \midrule
    \multicolumn{5}{l}{\textit{Rate-constrained}} \\
    \midrule
    Standalone & \textbf{206} & \textbf{1.67} & \textbf{2.25} & \textbf{4.31} \\
    GDeflate & 6,042 & 49.17 & 45.06 & 105.48 \\
    Zstd & 3,124 & 25.42 & 44.84 & 76.08 \\
    \midrule
    \multicolumn{5}{l}{\textit{Unconstrained}} \\
    \midrule
    Raw & 12,288 & 100.00 & 0 & 122.88 \\
    GDeflate & 14,327 & 116.59 & 45,38 & 188.65 \\
    Zstd & 12,727 & 103.57 & 46.02 & 173.29 \\
    \bottomrule
\end{tabular}
\end{table}

\begin{table}[!t]
\centering
\caption{CPU codec performance}
\label{table:speed-cpu}
\begin{tabular}{lrrrr}
    \toprule
    Codec & Rate (BPT) & Rate (\%) & Time ($\mu$SPT) & W2W ($\mu$SPT) \\
    \midrule
    \multicolumn{5}{l}{\textit{Rate-constrained}} \\
    \midrule
    Standalone & \textbf{206} & \textbf{1.67} & 26.16 & \textbf{28.22} \\
    Deflate & 1,701 & 13.84 & 13.69 & 30.7 \\
    Zstd & 2,231 & 18.16 & \textbf{12.19} & 34.5 \\
    \midrule
    \multicolumn{5}{l}{\textit{Unconstrained}} \\
    \midrule
    Raw & 12,288 & 100.00 & 0 & 122.88 \\
    Deflate & 11,938 & 97.15 & 13.36 & 132.74 \\
    Zstd & 11,852 & 96.45 & 12.28 & 130.80 \\
    \bottomrule
\end{tabular}
\end{table}

Table~\ref{table:speed-gpu} and Table~\ref{table:speed-cpu} show the codec performance results for GPUs and CPUs, respectively. In the \textit{Rate-constrained} set of experiments, the baseline methods use the rate-constrained representations from the standalone method. The \textit{Unconstrained} methods correspond to the compression of quantized representations from a vanilla GPT-2 model with no rate constraints. The \textit{Raw} method transmits an uncompressed representation at 16 bits per value, since the model uses the \textit{bfloat16} floating-point format. For GPU benchmarks, we compare with GDeflate \cite{GDeflate}, a variant of Deflate optimized for GPUs. The coding time measurements are reported in microseconds per token ($\mu$SPT) and the wall-to-wall (W2W) measurements assume an effective link speed of 100 Mbps and no protocol overhead.

The rate of the standalone codec is substantially lower than the benchmarks. Moreover, compressing the rate-constrained representations produced by our analysis transforms using off-the-shelf lossless codecs seems to have significant advantages in both rate and time over unconstrained representations. This result shows that even if we opt to use an off-the-shelf codec, there is still a significant benefit in using the rate-constrained representations induced by our methods. 

In GPU benchmarks, the standalone method outperforms the traditional lossless codecs in both rate and time. We exploited the high-parallelization capability of the standalone entropy models to achieve a significant time advantage. On the CPU, Zstandard, which uses the same dictionary-matching algorithm as DEFLATE, produces higher rates but is considerably faster than the other methods. The inference time of the entropy model on the GPU is 4.41\% of the total coding time. This places most of the current time overhead on the entropy codec. Using \textit{tabled ANS} (tANS) \cite{Duda2013AsymmetricNS}, a variant of an asymmetric numeral system (ANS), could help close the time performance gap, as it is the entropy coder used in the faster Zstandard method \cite{Collet2018ZstandardCA}. However, this change would most likely result in higher memory consumption. Nevertheless, compared to the Zstandard codec applied to our rate-constrained representations, at the current coding speed, and assuming a communication protocol overhead of 9\% \cite{Cavanaugh1994ProtocolOI}, the standalone codec is more efficient when the effective link speed is less than 153.07 Mbps.

\subsection{Comparative methods for the analysis of the rate-distortion performance and the split point}

The language modeling tasks use the same dataset and settings as discussed in \cref*{sec:rd-benchmarks}. The image classification tasks use ImageNet-1k \cite{RussakovskyDSKS15} and the corresponding settings are similar to those well-established in TorchVision \cite{torchvision2016}, including data augmentation, loss functions, and training methods, their parameters and schedules. For the ResNet method, we evaluate split points at blocks 3, 7, and 13, since the model is deeper.

To adapt our entropy model to a ResNet, a convolutional layer is prepended to the entropy model so, across all split points, the resulting number of channels is $E=768$ and the spatial dimensions are $7 \times 7$. These dimensions are flattened out and used as input for the entropy model after learned positional embeddings have been added. A dense layer is appended to the entropy model to recover the initial dimensionality.

The $\lambda$ hyper-parameter is chosen so that the distortion obtained at subsequent split points is as close and higher than previous points. Thus, the changes in rate cannot be attributed to different distortions. 

\subsection{Lipschitz constant of the optimized entropy model}

\begin{figure*}[!t]
    \subfloat[Split point]{
        \includegraphics[width=0.485\linewidth]{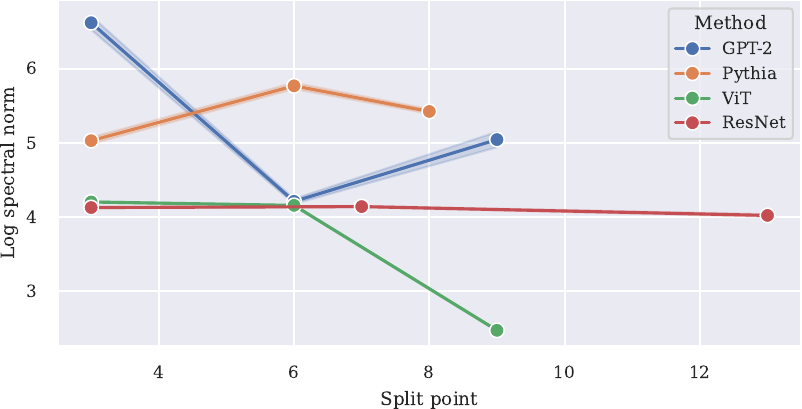}
        \label{fig:lip}
    }
    \subfloat[Bitrate]{
        \includegraphics[width=0.485\linewidth]{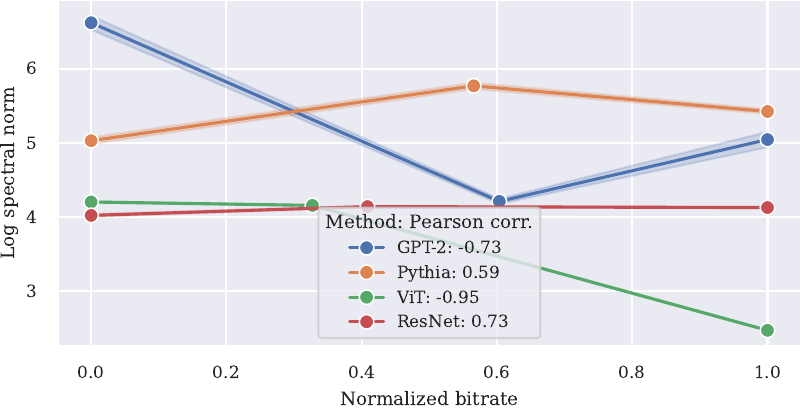}
        \label{fig:lip-bpt}
    }
    \caption{Estimates of the Lipschitz constant at different split points and corresponding bitrates, for GPT-2 Small, Pythia 160M, ViT B/16 and ResNet 34. The logarithmic scale is used. Bands show the standard deviation of the estimates over 10 samples.}
\end{figure*}

The Lipschitz constant of $r \in \mathcal{V}_r$ is approximated using the power iteration method on its Jacobian:
\begin{align}
    \bar{\mathfrak{L}}(r)
    =
    \frac{1}{N}
    \sum_{\mathbf{y} \in \mathcal{D}}
    \sqrt{
    b^\top_K
    J^\top_r
    (
    \mathbf{y}
    )
    J_r
    (
    \mathbf{y}
    )
    b_K
    }
    ;
    \displaystyle
    \quad
    b_{k+1}
    =
    \frac{
    J^\top_r
    (
    \mathbf{y}
    )
    J_r
    (
    \mathbf{y}
    )
    b_k
    }{
    \norm{
    J^\top_r
    (
    \mathbf{y}
    )
    J_r
    (
    \mathbf{y}
    )
    b_k
    }_2
    }
    ,
\end{align}
where $b_0$ is initialized randomly such that $\norm{b_0}_2 = 1$. Fig.~\ref{fig:lip} shows these measures per split point, for $N=100$, $K=1,000$, and $T=512$.

Fig.~\ref{fig:lip-bpt} shows the correlation between the Lipschitz constant of the trained entropy models and the bitrate of their target representations, for each of the codecs previously evaluated. With perhaps the exception of the ViT method, the correlation between these quantities seems rather weak.
\end{document}